\documentclass[a4paper,fleqn]{cas-dc}

\ExplSyntaxOn
\cs_gset:Npn \__first_footerline:
  { \group_begin: \small \sffamily \__short_authors: \group_end: }
\ExplSyntaxOff
\usepackage[authoryear]{natbib}
\setcitestyle{maxcitenames=2}
\usepackage{subfiles}
\usepackage{enumitem}
\usepackage{subfiles} 
\usepackage{xspace} % use to identify new command
\usepackage{amsmath} 
\usepackage{amsfonts}
\usepackage{adjustbox}
\usepackage{graphicx}
\usepackage{lipsum}
\usepackage{mathrsfs}
\usepackage{booktabs}
\usepackage{xr}
\usepackage{booktabs,multirow}
\usepackage{graphicx}
\usepackage{subcaption} 
\usepackage{algorithm}
\usepackage{algpseudocode}
\usepackage{placeins}
\usepackage{float}
\hypersetup{
    citecolor=cyan,
    linkcolor=cyan% change this to any color you want
}
\newcommand{\graph}{\mathcal{G}} % for all graph
\newcommand{\nodeSet}{\mathcal{V}} %for all nodes set
 
\newcommand{\centroidSet}{\mathcal{O}}
\newcommand{\centroid}{o}

\newcommand{\edges}{\mathcal{E}}
\newcommand{\relation}{\mathcal{R}}

\newcommand{\class}{\mathcal{C}} % set of class
\newcommand{\inputs}{\mathcal{X}}

\newcommand{\embeddingSet}{\mathcal{H}} % embedding set
\newcommand{\embedding}{\mathbf{h}} % individule embedding
\newcommand{\labelEmbedding}{\mathbf{h}^{label}}
\newcommand{\GNNpre}{\textit{GNN}^{pre}} %GNN in pre-process
\newcommand{\GNNinit}{\textit{GNN}^{init}} %GNN for label nodes 
\newcommand{\GNNref}{\textit{GNN}^{ref}} %first GNN
\newcommand{\parameter}{\theta} % parameters for basic GNN

\newcommand{\logits}{\mathbf{z}}

\newcommand{\loss}{\mathcal{L}}
\newcommand{\lossCE}{\mathcal{L}^{ce}}

\newcommand{\lossOT}{\mathcal{L}^{orth}}
\newcommand{\ourmethod}{\text{CTP}}

\def\tsc#1{\csdef{#1}{\textsc{\lowercase{#1}}\xspace}}
\tsc{WGM}
\tsc{QE}
\begin{document}
\let\WriteBookmarks\relax
\def\floatpagepagefraction{1}
\def\textpagefraction{.001}

\shorttitle{}    

\shortauthors{}  

\title [mode = title]{A Cross-graph Tuning-free GNN Prompting Framework}  




\author[1]{Yaqi Chen}

\author[1]{Shixun Huang}
\cormark[1]
\ead{shixunh@uow.edu.au}

\author[1]{Ryan Twemlow}
\author[1]{Lei Wang}
\author[1]{John Le}
\author[2]{Sheng Wang}
\author[1]{Willy Susilo}
\author[1]{Jun Yan}
\author[1]{Jun Shen}

\cortext[1]{Corresponding author}

\affiliation[1]{organization={University of Wollongong},
            addressline={Northfields Avenue}, 
            city={Gwynneville},
            postcode={2500}, 
            state={NSW},
            country={Australia}}

\affiliation[2]{organization={Wuhan University},
            addressline={No. 299, Luojia Hill}, 
            city={Wuhan},
%          citysep={}, % Uncomment if no comma needed between city and postcode
            postcode={430072}, 
            state={Hubei},
            country={China}}

% \ead{wsusilo@uow.edu.au}


\begin{abstract}
GNN prompting aims to adapt models across tasks and graphs without requiring extensive retraining. However, most existing graph prompt methods still require task-specific parameter updates and face the issue of generalizing across graphs, limiting their performance and undermining the core promise of prompting.
In this work, we introduce a \textbf{C}ross-graph \textbf{T}uning-free \textbf{P}rompting Framework (\textbf{\ourmethod}), which supports both homogeneous and heterogeneous graphs, can be directly deployed to unseen graphs without further parameter tuning, and thus enables a plug-and-play GNN inference engine.
Extensive experiments on few-shot prediction tasks show that, compared to SOTAs, \ourmethod{} achieves an average accuracy gain of 30.8\% and a maximum gain of 54\%, confirming its effectiveness and offering a new perspective on graph prompt learning. 
\end{abstract}


% \begin{highlights}
% \item 
% \item 
% \item 
% \end{highlights}

%\nocite{*}

% Keywords

\maketitle

% Main text
% \section{}\label{}
% \documentclass[main_anonymous.tex]{subfiles}   % 路径指向主文件
% \begin{document}

\section{Introduction}
% {\color{red} all comments (address comments (1) to (6) and some comments left across the paper by our next Tuesday meeting, make sure it is a read-for-submission version): (1) change the font of \textbackslash paragraph to the same format as AAAI. (2) add more references, make them to have one full page. (3) shorten some description to avoid there is only a few words in a sentence/line. (4) remember to remove mention 'refer to appendix' if it is moved to the main paper. (5) about moving from appendix to the main paper, priprotize the method figure, and experiment figures/results. (6) add more experiment result observation/insights and make sure that figures are close to the main text. do not separate them far away. (7) optimize some experiment performance number for our method (just leave it here as a memo first).(8) think about rephrase abstract and title (just leave it here as a memo first). }

% first point out the limitation of current model need train to apply on different dataset or task so it will be costy
% {\color{red} can consider adding many references in this paragraph about GNN.}
Graph Neural Networks (GNNs) possess strong relational modeling capabilities, supporting applications that span across social networks, knowledge bases, biological systems, and recommender engines \citep{zhou2020graph,graphsage,zeng2020graphsaint,he2020lightgcn}.
%\citep{wu2020comprehensive,zhou2020graph,graphsage,zeng2019graphsaint,vashishth2019composition,GAT,he2020lightgcn}. 
 However, their practical deployment remains limited. Most existing GNNs are developed under a graph-specific setting, where a model is trained and evaluated on different splits of the same graph rather than transferred to unseen graphs. This setting is often unrealistic in practice, where downstream graphs may vary substantially in structure, semantics, and even graph type. In addition, many existing approaches depend on supervised signals or task-specific labels during training, which reduces their scalability and limits their applicability in label-scarce real-world scenarios. Even when transferred to a new graph, these methods often require additional fine-tuning or downstream adaptation, resulting in considerable computational cost and reduced usability. Therefore, a practically useful graph learning framework should not only generalize across graphs, but also reduce dependence on labeled data during training and avoid task-specific re-optimization during deployment.

% [current studies] To fix this several study have explore 3 kind of methods: 1.pre-train and fine tune, 2.tune-dependent prompt, 3.non-tunable prompt. 
% result the problem setting gap: not much study about non-tunable prompt

\begin{figure} 
    \centering
    \includegraphics[width=0.5\textwidth]{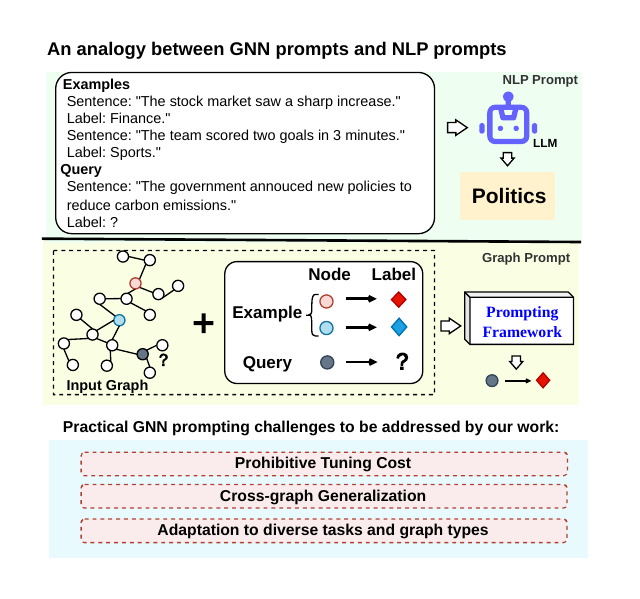}
    \caption{An analogy between GNN and NLP prompts}
    \label{fig:intro}
\end{figure}
%\vspace{-3ex} % 负值减小间距，自己调节

To overcome these limitations, existing studies have mainly followed two paradigms: \emph{pre-train and fine-tune} and \emph{prompt learning}. The \emph{pre-train and fine-tune} paradigm leverages large-scale graph information to learn transferable representations, which are then adapted to downstream tasks through task-specific fine-tuning \citep{hu2020gpt, hu2019strategies, xie2022self, subramonian2021motif}. Although fine-tuning avoids training from scratch, it still requires updating GNN parameters for new tasks or graphs, limiting scalability. In contrast, \emph{prompt learning}, inspired by NLP prompts that use task descriptions and exemplar outputs to guide model inference, extends this paradigm to the graph domain (Figure~\ref{fig:intro}). Since graphs lack the sequential semantics that naturally guide prompt design in text, researchers resort to introducing small learnable components, called graph prompts, attached to a frozen GNN backbone to encode task-specific signals~\citep{graphprompt,allinone,gppt,universal,li2024adaptergnn}.

These tuning-dependent prompts are typically implemented as trainable tokens or answer-function modules. By tuning their parameters, the prompts either adapt a frozen GNN backbone to downstream objectives or transform unseen inputs to better align with the pretrained model. Although this strategy substantially alleviates the heavy cost of full-model fine-tuning, it still relies on downstream parameter updates and therefore falls short of truly plug-and-play deployment. Moreover, many existing graph prompting methods are trained or adapted with task-specific supervision, and are often evaluated within the same graph or a single graph type, leaving their self-supervised capability and cross-graph generalization largely underexplored.

%Notably, a truly \emph{tuning‑free prompt}—one without task‑specific optimization—remains largely unexplored, exposing three gaps. 

\noindent \textbf{The Research Gaps}.  Despite the growing progress of graph prompt learning, existing methods still fall short of supporting a truly general and deployment-ready prompting paradigm. 
We summarize the main research gaps as follows. 
\begin{itemize}[leftmargin=*]
    \item \emph{Gap 1: Over-reliance on prompt tuning.} Unlike conventional NLP prompts ~\citep{brown2020language,liu2023pre}, which function as an interface between the task and the pretrained model rather than as an additional trainable module, many existing graph prompting methods~\citep{graphprompt,gppt, universal,allinone} deviate from this original motivation. Instead of using prompts purely as task reformulation, they often introduce trainable prompt tokens, prompt structures, or answer-function modules that must be optimized for each downstream task. As a result, these approaches increases training cost and reduces the efficiency and simplicity advantages of prompting, and falls short of the truly plug-and-play usage that prompting is intended to enable.
    
    \item  \emph{Gap 2: Over-reliance on task-specific supervised signals}. Some existing graph prompting approaches~\citep{wu2020comprehensive,zhou2020graph,graphsage,zeng2020graphsaint,vashishth2019composition,GAT,he2020lightgcn} depend on labeled signals (e.g., node labels), either to optimize prompts directly or to adapt prompt-related components for specific tasks (e.g., node classification). Such reliance on supervision weakens their applicability in real-world graph domains, where high-quality labels are often scarce, expensive, or unavailable. More importantly, it leaves open the question of whether graph prompting can be developed in a fully self-supervised manner while retaining strong transferability across tasks. 
    
    \item \emph{Gap 3: Transductive limitation}. Many prior studies~\citep{graphprompt,yang2024graphpro,gppt} assess graph prompting in a same-graph setting, where training and evaluation are conducted on different splits of a single graph. Although this protocol is convenient for benchmarking, it does not reflect realistic deployment scenarios, where the pretrained model is expected to generalize inductively to entirely unseen graphs. %As a result, the cross-graph generalization ability of existing graph prompting methods remains insufficiently explored. 

    \item \emph{Gap 4: Limited support for diverse graph types}. Current graph prompting methods~\citep{yu2024hgprompt,allinone} are typically developed for only one graph type, such as homogeneous graphs or knowledge graphs, rather than a unified setting covering both homogeneous and heterogeneous graphs. This limitation restricts their applicability in real-world scenarios where graph data exhibit diverse structural and semantic properties.
\end{itemize}

%Taken together, these limitations suggest that existing GNN prompt learning has not yet fully addressed the need for a self-supervised, tuning-free, and cross-graph transferable framework that can generalize across both unseen graphs and different graph types.

\noindent \textbf{Our goal}. To address the above limitations, we design a \emph{self-supervised}, \emph{tuning-free}, and \emph{inductive} GNN prompting framework that generalizes across both \emph{homogeneous} and \emph{heterogeneous} graphs for fundamental tasks, including node classification and link prediction. Inspired by the in-context meta learning paradigm~\citep{dong2022survey,min2021metaicl,prodigy}, we leverage an unsupervised prompting framework which unifies the pretraining task and downstream tasks into one single task. This unified representation enables a single model to handle diverse graph tasks without additional tuning, regardless of graph size. In this framework, a prompt graph is first constructed to contextualize input nodes and edges by incorporating both prompt examples and query instances, which are connected through label nodes to establish direct prompt–query interactions. A GNN is then pretrained on these prompt graphs, where attention mechanisms facilitate information propagation, enabling strong out-of-the-box performance across downstream tasks. The in-context learning objective includes a self-supervised neighbor-matching task that predicts the neighborhood structure of each node or edge, thereby promoting effective relational reasoning.
In summary, we make the following contributions:

%To the best of our knowledge, PRODIGY is the only tuning-free prompt framework for GNNs \cite{prodigy}.
%Inspired by the in-context learning paradigm \cite{dong2022survey}, it defines each input’s neighborhood as its context, collects prompt examples along with their contexts, and generate the same label for example belonging to the same neighborhood. The examples and labels are then assembled into a prompt graph, which models the relationships between them and enables a frozen GNN to perform downstream tasks without parameter updates.

%Although the framework is tuning-free, several issues remain unresolved from the pespective of \emph{sampling, context construction and training objective}: First, the purely random sampling causes unbalanced graph coverage. It tends to oversample nodes in dense regions while undersampling nodes in sparse regions, leading to uneven sample distribution. Second, the context construction can introduce inconsistent signals within the same context, where contradictory examples mislead the model; third, 
%there is still much room for the objective design such that the node and label embeddings can be more distinguishable for classification tasks.

%These methodological gaps motivate us to propose a tuning-free prompt framework that address those gaps, yielding robust performance across both homogeneous and heterogeneous graphs and supports node- and edge-level tasks. 

\begin{itemize}

\item We introduce our \textbf{Cross-graph Tuning-free Prompting} Framework (\textbf{\ourmethod}) that addresses key research gaps in GNN prompt learning. It is trained in a self-supervised manner and can be directly deployed to unseen graphs and tasks without parameter tuning.
\item In this framework, we propose a self-supervised prompt sampling strategy, a structure-aware contextualization mechanism that jointly models prompt examples and query instances by leveraging multiple structural properties to enhance inductive generalization, and an effective training objective for capturing diverse relational patterns and improving task flexibility.

%To bridge the methodology gaps, we introduce multiple non-trivial improvements covering the major steps of the whole framework, such as a self-supervised sampling process for representative prompt construction, a context construction method for balancing diversity and integrity, and an optimization of the training objective to enhance label separation and classification accuracy.
\item Extensive experiments demonstrate that our framework consistently outperforms baseline methods on few-shot prediction tasks across both homogeneous and heterogeneous graphs. Compared with state-of-the-art approaches, it achieves an average improvement of 30.8\% and a maximum gain of 54\%, while exhibiting the relatively low performance variance.
\end{itemize}

\section{Related Work}
\label{related}
% Start with pre-train and fine-tune, first the GNN still need to be fine-tune later prompt been introduced, the GNN is frozen after pre-training, fine-tuning prompt to adapt to downstream task. From setting perspective, first use same dataset but different task later use inductive setting only PRODIGY use totally different dataset.
% {\color{red} can consider introduce more description and referenes here (e.g., another paragraph about GNN). but this is the last one to do considering the page size.}
\noindent \textbf{Pre-training and fine-tuning. }To alleviate label scarcity and improve generalization, graph models commonly adopt a pre-train fine-tune paradigm: a self-supervised pre-training stage captures transferable structural regularities, followed by light supervision on the downstream task. Typical pretext families include autoencoding (e.g., marginalized graph autoencoders)~\citep{wang2017mgae}, masked/component modeling and multi-level strategies that jointly pre-train node- and graph-level objectives~\citep{hu2019strategies}, and graph contrastive learning at node/graph levels (e.g., DGI~\citep{deepgraph}, InfoGraph~\citep{sun2019infograph}, GraphCL~\citep{contrastive}). Large-scale molecular pretraining with graph Transformers (e.g., GROVER~\citep{GROVER}) further shows the benefit of rich self-supervision before task-specific fine-tuning. While effective, this pipeline still updates backbone parameters per task/domain and can struggle when task distributions diverge from pretext objectives.

\smallskip
\noindent \textbf{Meta in-context learning.}
Meta in-context learning (meta-ICL) provides a \textit{tuning-free} route to downstream adaptation: the model is meta-trained to adapt in the forward pass from few-shot prompts, so test-time weight updates are unnecessary~\citep{min2021metaicl}. Prior work shows (i) large language models already perform few- or zero-shot ICL by conditioning on exemplars~\citep{brown2020language,li2024meta}, and meta-training further strengthens such behavior across diverse tasks~\citep{min2021metaicl}; and (ii) ICL can be interpreted as gradient descent executed inside the forward computation~\citep{von2023transformers, akyurek2022learning} or as implicit Bayesian updating driven by coherent task priors~\citep{xie2021explanation}. In GNN, related meta-learning ideas have been explored for cross-graph few-shot generalization~\citep{huang2020graph,zhou2019meta}.
%(e.g., G-Meta uses local subgraphs as transferable units)

\smallskip
\noindent\textbf{Graph Prompt Learning}.
Instead of fully fine-tuning all model weights~\citep{surveyGrpahPrompt}, graph prompt learning aims to bridge the gap between pretraining and downstream objectives by conditioning a pretrained GNN on task-relevant prompt signals. Early prompt-based graph methods still required updating substantial portions of the GNN backbone, making them more akin to lightweight fine-tuning rather than true prompting~\citep{hu2019strategies, GROVER}. Later works moved toward \emph{prompt-tuning}, where the backbone is frozen and only prompt-related parameters are optimized. 
For example, GPF~\citep{hu2020gpt} learns universal prompt features appended to node inputs, while GraphPrompt~\citep{graphprompt} unify pretraining and downstream tasks through a shared template and inject learnable prompts at the readout layer or throughout the encoder. GraphPro~\citep{yang2024graphpro} combines structural prompts for recommendation, HGPrompt~\citep{yu2024hgprompt} further extends this line to both homogeneous and heterogeneous graphs via a dual-template design, and EdgePrompt~\citep{fu2025edge} explores a different design choice by learning edge prompts and incorporating them through message passing. More recently, LEAP~\citep{xu2025learning} strengthens the universality perspective of graph prompt tuning and uses actor-critic reinforcement learning to select nodes and edit prompts for better generalization. However, despite their different prompt designs, these methods still require prompt optimization on downstream tasks, and thus remain fundamentally tuning-dependent.

Another line of work incorporates meta-learning into graph prompting to improve prompt initialization and adaptation. All-In-One~\citep{allinone} reformulates node-, edge-, and graph-level tasks into a unified graph-level prompting format and meta-learns prompt initialization across tasks, improving transferability but still requiring per-task prompt tuning. PRODIGY~\citep{prodigy} takes a more decisive step toward tuning-free graph in-context learning by introducing an explicit prompt graph that connects support examples, queries, and label nodes, and by meta-training a GNN to solve downstream tasks in context without any parameter updates at test time. To the best of our knowledge, it is the first graph framework to clearly realize the tuning-free prompting paradigm on unseen graphs, demonstrating effective transfer across both citation networks and knowledge graphs. However, the current framework does not fully exploit its potential. Random sampling and augmentation may introduce noise and distort structural patterns, while weak modeling of inter-label relationships can blur class boundaries. This leaves room for improving how information is organized and reused across tasks. GraphPrompter~\citep{lv2025graphprompter} extends the PRODIGY-style framework with multi-stage prompt generation, selection, and augmentation, reducing noisy prompts and improving accuracy on diverse downstream graphs. However, its performance gains stem from more sophisticated learned optimisation components rather than from eliminating the dependence on supervised training signals. In contrast, our goal is not only to improve graph in-context performance, but also to establish a framework that is fully self-supervised during pretraining, tuning-free at deployment, and applicable across different tasks and graph types under a unified formulation.

Beyond prompting, recent work on fully-inductive graph generalization also highlights the importance of transferring to unseen graphs. GraphAny~\citep{zhao2024fully} defines a fully-inductive setting in which test graphs may have new structures, feature spaces, and label spaces, and proposes an analytical inference scheme with learned inductive attention for arbitrary graphs. This setting is closely related to our cross-graph motivation. However, GraphAny is not a self-supervised prompting framework; it learns from labeled nodes and focuses specifically on node classification rather than unified prompting across graph types and task forms.

Overall, despite recent progress, several notable limitations remain. Most existing approaches still require downstream prompt optimisation, many rely on supervised adaptation rather than fully self-supervised training, and support across diverse graph types is fragmented. These challenges motivate our work, which aims to address the unresolved gaps in prior studies.

\begin{figure*}[!h] 
    \centering
    \includegraphics[width=\textwidth]{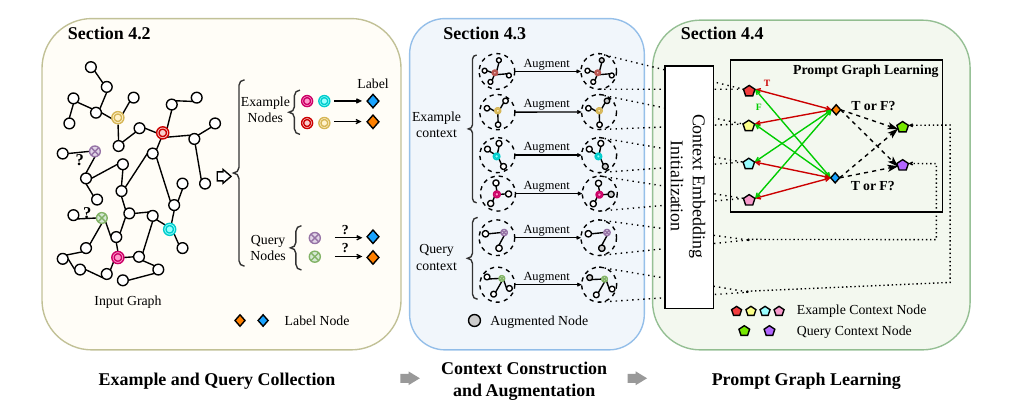}
    \caption{
        \textbf{The Prompting Framework.}
        The overall workflow includes three stages, illustrated with node classification. During example and query collection, example and query nodes are sampled for each class. Then each sampled node’s context is constructed and augmented from its neighborhood. Finally a prompt graph is built, with context nodes initialized using their contextual embeddings. A learning process refines all node embeddings and all learned parameters will be directly used on unseen test graphs without any further update.
        %\shixun{first component: to do: context embedding initliazation. second one: prompt graph learning. let the second component include all the promt graph. add 'input graph' to the input graph. add notations to 'input graph' and say sth like 'input graphs are different during pretraining and testing phases'. for two gnn componoenents, give notations like 'the pretrained parameters will be directly applied in the testing phase without any fine tuning.'}
        }
    \label{fig:prompt_pipeline}
\end{figure*}
% \documentclass[main_anonymous.tex]{subfiles}   % 路径指向主文件
% \begin{document}
\section{Preliminaries}

\noindent \textbf{Unified Formulation of Node and Link Classification}.
Let $\graph = (\nodeSet, \edges, \relation)$ be a graph, where each edge $e = (u, r, v) \in \edges$ connects two nodes $u, v \in \nodeSet$ via relation $r \in \relation$. Let $\class$ denote a set of $m$ classes. We treat node classification and link prediction as a unified task. Specifically, depending on the task, we define $\inputs$ as a set of individual nodes or node pairs from $\nodeSet$. For each $x \in \inputs$, $x$ contains a single node (i.e., $|x| = 1$) for node classification, whereas for link prediction, $x$ contains the two endpoints of a potential edge (i.e., $|x| = 2$). Each input $x$ is assigned with a label $c\in \class$ and each sample is then represented as a tuple $(x, c)$.
%not example

%\shixun{you still have not introduced the definition of $\inputs$,e.g., $\inputs=?$ and ? can be a set of single node or a pair of nodes depending on the task; why $b$ and $d$ instead of $u$ and $v$?; shall we mention the last part in the following paragraph? }

\smallskip
\noindent \textbf{Inductive, Few-shot, Tuning-free Prompting}.
For an $m$-way, $s$-shot episode, we construct a support set $\mathcal{S}^{ex} = \{(x_i, c_i)\}_{i=1}^{s \cdot m}$ containing exactly $s$ labeled examples per class, and a query set $\mathcal{S}^{qu} = \{(x_j, c_j)\}_{j=1}^{n \cdot m}$ with $n$ queries per class. We focus on the inductive setting in this paper, the pretrained model in the source graph will be directly applied to a different downstream graph without updating any learnable parameters (including prompts). 

%\shixun{comment:  'inductive' setting is not well explained; you should use c instead of y in S; y seems to be useless. we need label, not label embedding!}
%change y to c,(b,d) become (u,v)-done
% \iffalse

% In downstream

%\textbf{Motivation of this paper}Despite the effectiveness of this design, there remains significant room for improvement across its core components. In the following sections, we analyze the limitations of standard techniques used in each component and present our proposed optimizations.

% \end{document}
% \documentclass[main_anonymous.tex]{subfiles}   % 路径指向主文件
% \begin{document}
\section{Methodology}
\label{method}

In this section, we first provide an overview of the prompt learning framework (inspired by~\cite{prodigy}) in Section~\ref{sec:framework}, followed by our key optimisations within its core components in Sections~\ref{sec:collection}–\ref{sec:learning}.

\subsection{The Prompt Learning Framework Overview}\label{sec:framework}

Our framework (illustrated in Figure~\ref{fig:prompt_pipeline}) comprises three fundamental components, each described in detail below.

\begin{itemize}
    \item \emph{Example and Query Collection}.
% During pretraining highlight it's few-shot and self-supervised
During pre-training, the process begins by sampling a set of neighborhoods, with each neighborhood representing a pseudo class (i.e., a self-supervised signal). From each neighborhood, a small number of examples and queries (belonging to the same class) are sampled to construct the example set and query set. In the testing phase, we have an entirely new input graph and the collected examples and queries have external class labels (e.g., node category labels and edge type labels). When the downstream task is edge type classification, each example or query contains two nodes.

\item \emph{Context Construction and Augmentation}.
% identify context
Although example and query nodes are defined, each node alone lacks sufficient context for effective learning. To address this, local neighborhoods are used to provide contextual information.
To mitigate the issue of limited information, an augmentation step inspired by contrastive learning is applied during pre‑training. It corrupts the context of example and query nodes via \emph{node dropping} and \emph{feature masking}, enhancing robustness to structural and attribute variations.

\item \emph{Prompt Graph Learning via Edge Type Prediction}.
%how prompt graph constructed
To enable tuning-free adaptation, a prompt graph been introduced. It contains context nodes (examples and queries) initialized via $\GNNinit$ using neighborhood aggregation, and label nodes initialized by standard methods. A separate model $\GNNref$ refines all embeddings. The prompt graph formulates classification as an \emph{edge type prediction} task: an edge between a context node and a label node is labeled True if they belong to the same class. Both GNNs are trained only during pretraining and reused directly during testing.

\end{itemize}

%\paragraph{Motivation of this paper}Despite the effectiveness of this design, there remains significant room for improvement across its core components. In the following sections, we analyze the limitations of standard techniques used in each component and present our proposed optimizations.

%To enhance the overall performance of the framework, we propose new techniques targeting each of the three core components. In this section, we detail these improvements individually, providing both the motivation and the corresponding methodology.

\begin{figure*}[!htbp]
    \centering
    \includegraphics[width=\textwidth]{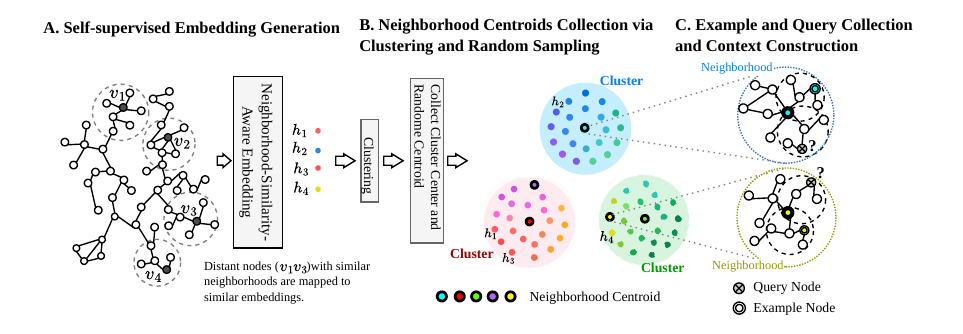}
    \caption{
        \textbf{Neighborhood Centroid Collection.} This figure illustrates the preprocessing step for clustering node embeddings and selecting centroids. (A) Initially, node embeddings are generated by a self-supervised GNN, enabling them to capture neighborhood similarity in the graph. (B) The embeddings are then clustered, and the cluster center nodes together with some randomly sampled nodes are collected to form the neighborhood centroid candidate set. (C) Each centroid represents a class, and example and query nodes are subsequently sampled from the local neighborhood of each centroid for context construction. Unless otherwise specified, all centroids in the following sections refer to neighborhood centroids. }  
    \label{fig:technique 1}
\end{figure*}

\subsection{Representative Example and Query Collection for Pretraining}
\label{sec:collection}

% motivation of reverse this part
In the self‑supervised setting, labels are assigned by neighborhood, with example and query nodes in the same neighborhood sharing a label. The selection of neighborhoods is determined by the neighborhood centroids being sampled. Unless otherwise specified, all centroids in the following sections refer to neighborhood centroids. Centroid sampling is therefore crucial, as the chosen centroids should ideally cover all classes. However, without access to true labels, direct sampling is infeasible. A common workaround is random centroid sampling~\citep{prodigy}. But, due to the non‑uniform degree distributions of graphs, this method tends to oversample nodes from dense regions, leading to under‑representation of small classes in sparse regions and loss of valuable samples.

To address this issue, we introduce a collection strategy driven by neighborhood similarity. As a preprocessing step, we first generate node embeddings using a self‑supervised GNN, which capture neighborhood similarity. That is, nodes with similar local structures obtain similar representations even if they are far apart in the graph. We then cluster these embeddings to form similarity‑aware groups and select cluster centers as centroid candidates, reducing the risk of missing rare patterns. Figure~\ref{fig:technique 1} illustrates our approach.

We use a spatial $\GNNpre$ for local aggregation to capture low-order topological information with a random-walk based objective function to capture higher-order information. The result embedding set $\embeddingSet^{pre}$ contains embeddings (with $d$ dimensions) of all nodes from the input graph $\graph^{input}$:
% {\color{red} we did not explain $\embeddingSet^{pre}$}:
\begin{equation}
    \embeddingSet^{pre} = \GNNpre(\graph^{input};\,\parameter^{pre}),
\end{equation}
\vspace{-1ex} % 负值减小间距，自己调节
\begin{equation}
\resizebox{\columnwidth}{!}{$
  \loss(\mathbf{h}^{pre}_u)
    = -\log\bigl(\sigma({\mathbf{h}^{pre}_u}^\mathrm{T}\mathbf{h}^{pre}_v)\bigr)
    - Q\,\mathbb{E}_{v_n\sim P_n(v)}\Bigl[\log\bigl(\sigma(-{\mathbf{h}^{pre}_u}^\mathrm{T}\mathbf{h}^{pre}_n)\bigr)\Bigr]
$},
\end{equation}

\noindent where $\GNNpre$ can be any spatial-based GNN (e.g., GraphSAGE), $\parameter^{pre}$ refer to learnable parameters, $\mathbf{h}^{pre}_u$ is the embedding of node $u$, node $v$ is a positive sample co-occuring in the same random walk as $i$, $Q$ defines the number of negative samples and negative sample node $n$ is drawn from the negative sampling distribution $P_n$. 

To leverage these embeddings, we cluster them (e.g., using $k$‑means) based on semantic and structural similarity. Instead of performing k-means for every batch using all available points, we perform a one-time k-means clustering with a substantially large k. Each batch then sample nodes from these precomputed clusters. This strategy captures finer intra-class variations and substructures, balancing granularity and computational efficiency.
 %\shixun{change the logic here. we need to explain what kinds of embeddings are expected and why self-supervised graphsage is a better choice. introduce the gnn objective function (random walk version).}

% Why only k-means is not enough
However, in sparse graph regions, pure clustering may still overlook rare patterns. Supplementing the clusters with randomly sampled nodes fills these coverage gaps and mitigates clustering bias, providing a more comprehensive training sample. To ensure representativeness and diversity, a centroid set $\centroidSet$ is constructed for all $b$ batches and used throughout training. Specifically, a centroid set consists of m-ways per batch, i.e., $\centroidSet=\{\centroid_i\}_{i=1}^{m\!\times\!b}$. It consists of cluster centers and randomly sampled nodes:
\begin{equation}
\resizebox{\linewidth}{!}{$
  \centroidSet
  =
  \operatorname{K-Means}\bigl(\embeddingSet^{pre}, k\bigr)
  \cup
  \operatorname{RandomSample}\bigl(\embeddingSet^{pre}, {|\centroidSet|\!-\!k}\bigr),
  $}
\end{equation}

\noindent where $k\!=\!\lfloor \alpha \cdot |\centroidSet|\rfloor$, and $\alpha \in [0,1]$. Here, $\operatorname{K-Means}$ runs on embeddings and returns the corresponding nodes, so $k$ cluster centers are obtained via $k$-means, while the remaining centroids are sampled uniformly at random.

For each centroid $\centroid$, we extract its $h$-hops neighborhood to obtain the centroid subgraph $\graph_\centroid = (\nodeSet_\centroid, \edges_\centroid, \relation_\centroid)$. Each $\graph_\centroid$ corresponds to a class $c \in \class$. From every $\graph_\centroid$, we uniformly sample $s$ example nodes to form the subset $\nodeSet^{ex}_o$ and $n$ query nodes to form the subset $\nodeSet^{qu}_o$. All nodes drawn from the same $\graph_\centroid$ share same label $c\in \class$ also align to the class. The example set ($\mathcal{S}^{ex}_o$) and query set ($\mathcal{S}^{qu}_o$) gathered from certain $\graph_\centroid$ to represent class $c$ can be represented as:
%emphesize o represent a class
\begin{equation}
\begin{aligned}
\mathcal{S}^{ex}_c 
  &= \bigl\{\,\bigl(x_{i},\,c\bigr)\,\big|\,x_i\in \nodeSet^{ex}_o,\;o \Leftrightarrow c\in \class\bigr\}, \\[4pt]
\mathcal{S}^{qu}_c 
  &= \bigl\{\,\bigl(x_{j},\,c\bigr)\,\big|\,x_j\in \nodeSet^{qu}_o,\;o\Leftrightarrow c\in \class\bigr\}.
\end{aligned}
\end{equation}

% \shixun{here we can add a recall of the meaning of n and s, since it is too far away from the definition.}
By incorporating the example-node subsets and query-node subsets from all subgraphs yields the example and query sets $\mathcal{S}^{\mathrm{ex}}$ and $\mathcal{S}^{qu}$, respectively. %\shixun{in equation 4, for $\mathcal{S}^{qu}_o$, it should be a tuple of two elements like $\mathcal{S}^{ex}_o$}

%\shixun{we need to show a figure, and describe how many samples and queries are sampled. perhaps for figure 1, nodes a,b,c and d. we show two hops?}Done this
% forming the example set $\exampleSet$ and the query set $\querySet$.

\subsection{Context Construction with Structural Diversity, Consistency and Integrity}
\label{sec:context}

\begin{figure*}[!tb]
    \centering
    \includegraphics[width=0.8\textwidth]{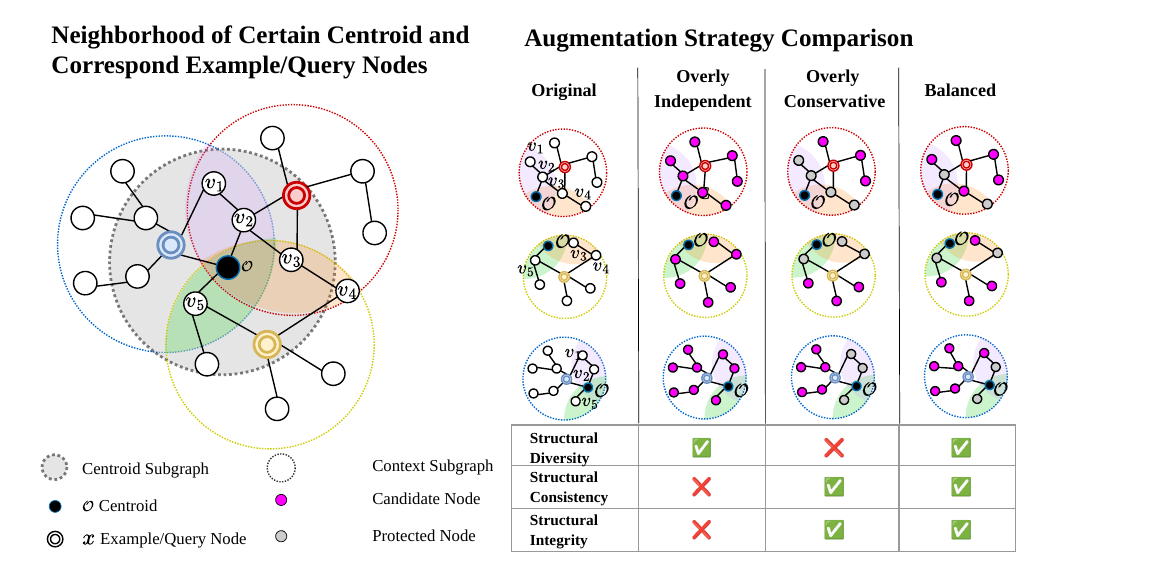}
    \caption{\textbf{Augmentation Strategy Comparison}.
        When sampling context near a shared centroid, subgraphs often overlap. The \emph{Overly Independent Approach} augments all nodes, including overlapping ones, which disrupts structural consistency and integrity. The \emph{Overly Conservative Approach} blocks augmentation in these regions, reducing structural diversity. Our \emph{Balanced Approach} selectively limits augmentation in overlapping areas, preserving consistency while ensuring sufficient diversity and integrity. }
    \label{fig:technique 2}
\end{figure*}

% how context constructed
For each input $x$, we extract its neighborhood within certain hops (specified in the parameter settings) to form a context graph. These subgraphs preserve the local structural and relational information around each selected node, serving as the input for pretraining.

% why we need augmentation (shortcome)
In few‑shot settings, the number of available context graphs is limited, which may cause the model to overfit specific features and hinder generalization. Data augmentation is commonly used to address this. In our case, context graphs are labeled based on samples from the same $\graph_\centroid$, which must remain small to avoid noise. Thus, context graphs with the same label exhibit significant overlap.

Under this setup, applying augmentation \emph{independently} to each context graph~\citep{prodigy}  would be a natural solution. This approach may produce samples with \emph{structural diversity}. However, it often compromises \emph{structural consistency} and \emph{structural integrity}. Regarding \emph{structural consistency}: due to the fact that nodes and patterns are shared across different context graphs, independent augmentations may modify shared nodes or pattern in some graphs but not others. As a result, augmentation misleads the model by introducing conflicting or corrupted information. For \emph{structural integrity}: since we rely on $\graph_\centroid$ to identify class semantics, and the overlapping regions are typically centered around the centroid itself, excessive augmentation in these areas may disrupt critical structural patterns and undermine the underlying semantic cues essential for classification. Thus, such an \emph{overly independent} approach can degrade performance by injecting spurious variations or disrupting critical structural patterns necessary for effective representation learning.

% introduce our method
To mitigate these issues, one might avoid augmentation in overlapping regions entirely. However, this \emph{overly conservative} approach is suboptimal. Overlapping regions often cover a large portion of context graphs; blocking them reduces augmentation diversity. %\shixun{\color{red} what does this mean?} It also risks biasing the model toward frequently appearing nodes, regardless of their true semantic relevance.

To strike a good balance among structural diversity, consistency and integrity, we propose a \emph{Balanced} augmentation strategy, illustrated in Figure~\ref{fig:technique 2}. Let $\nodeSet_\centroid$ denote all nodes in $\graph_\centroid$. The key idea is to protect a subset of these nodes from augmentation. We define:
\begin{equation}
\label{eq:nodeSet-centroid}
\nodeSet_\centroid \;=\; 
\{\centroid\} \;\cup\; \nodeSet^{ex}_\centroid \;\cup\; \nodeSet^{qu}_\centroid 
\;\cup\; \nodeSet^{remain}_\centroid,
\end{equation}
\noindent here, \(\nodeSet^{\mathrm{remain}}_\centroid\) denotes the set of nodes in \(\nodeSet_\centroid\) after removing the centroid \(\centroid\) as well as the sampled sets \(\nodeSet^{ex}_\centroid\) and \(\nodeSet^{qu}_\centroid\). Then we separate the nodes to be preserved into $\nodeSet^{\mathrm{protect}}_\centroid$:
\begin{equation}
\label{eq:nodeSet-split}
\begin{aligned}
    \nodeSet^{\mathrm{protect}}_\centroid
    &=\; 
    \{\centroid\} \;\cup\; \nodeSet^{ex}_\centroid \;\cup\; \nodeSet^{qu}_\centroid 
    \;\cup\; \nodeSet^{(p)},\\
\end{aligned}
\end{equation}
\noindent where $\nodeSet^{p} \subset \nodeSet^{remain}_\centroid$ contains a fraction $p$ of the nodes in $\nodeSet^{remain}_\centroid$. Nodes in $\nodeSet^{\mathrm{protect}}_o$ are exempt from augmentation. For every input $x$ from $\nodeSet^{ex}_\centroid \;\cup\; \nodeSet^{qu}_\centroid$, its context graph is defined as $\graph^{\mathrm{con}}_x = (\nodeSet^{\mathrm{con}}_x, \edges^{\mathrm{con}}_x, \relation^{\mathrm{con}}_x)$. A candidate node set $\nodeSet^{\mathrm{candidate}}_x \subset \nodeSet^{\mathrm{con}}_x$ is subject to augmentation operations and consists of all nodes in $\nodeSet^{\mathrm{con}}_x$ except those in $\nodeSet^{\mathrm{protect}}_\centroid$.
\begin{equation}
\nodeSet^{\mathrm{candidate}}_x=\nodeSet^{con}_{x}\ \;\setminus\; \nodeSet^{\mathrm{protect}}_\centroid.
\end{equation}

Then, two corruption methods, namely node dropping and feature masking, are applied to each context subgraph $\graph^{con}_{x}$. These methods are restricted to the node set $\nodeSet^{candidate}_x$ producing an augmented context subgraph $\widetilde{\graph^{con}_{x}}$.

\begin{figure*}[H]
    \centering
    \includegraphics[width=1\textwidth]{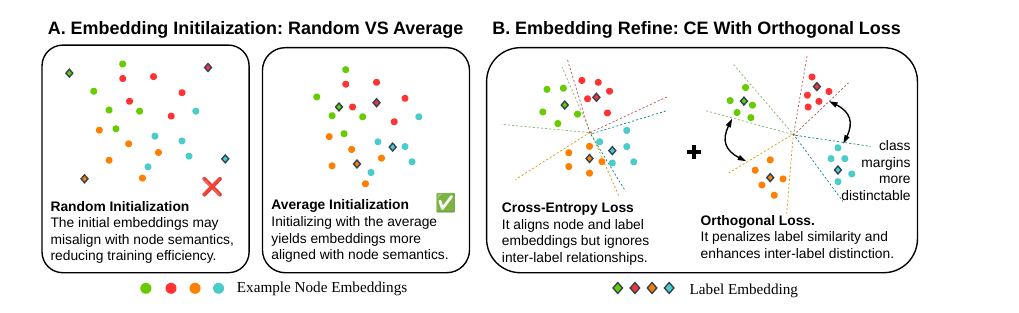}
    \caption{
        Average Initialization and Orthogonal Loss. 
        (A) Compared to random initialization, average embedding provides more context-related label representations. (B) Cross-entropy aligns labels with nodes but neglects inter-label relations, often leading to unclear boundaries. Orthogonal loss mitigates this by encouraging label separation. }
    \label{fig:technique3}
\end{figure*}
% Prompt graph initialization
With the augmented context‐graph $\widetilde{\graph^{con}_{x}}$ for each $x$, the first step in constructing the prompt graph is passing augmented context graphs through $\GNNinit$ (e.g., GCN~\citep{GCN} or GAT~\citep{GAT}) to generate node embeddings. Embeddings derived from context-graph for each $x$ in $\mathcal{S}^{ex}$ become example context nodes $\embedding^{ex}_x$, while those from $\mathcal{S}^{qu}$ become query context nodes $\embedding^{qu}_x$. 
\begin{equation}
\begin{aligned}
\embedding^{ex}_{x} &= \GNNinit(\widetilde{\graph^{\mathrm{con}}_{x}}), 
&& x \in \nodeSet^{\mathrm{ex}}, \\
\embedding^{qu}_{x} &= \GNNinit(\widetilde{\graph^{\mathrm{con}}_{x}}), 
&& x \in \nodeSet^{\mathrm{qu}}.
\end{aligned}
\end{equation}
% h need _x

For link–prediction task, the input $x$ contains two target nodes $v_1$ and $v_2$. To align the size of input, the neighbor nodes of each target are first gathered into a context subgraph, and embeddings $\embedding_i$ are generated for all nodes within this subgraph.
A second aggregation is then applied to the target nodes $v_1$, $v_2$ in $x$ to obtain $\embedding_{v_1}$ and $\embedding_{v_2}$.  
Max pooling across the entire subgraph yields $\embedding_{\max}$.  
These three embeddings are concatenated and passed through a linear projection layer to restore $d$ dimension.
\begin{equation}
\begin{split}
  \embedding_x
    = W^{\top}\!\bigl(\embedding_{v_1}\,\|\,\embedding_{v_2}\,\|\,\embedding_{\max}\bigr) + b,
  (v_1,\,v_2)\in \nodeSet^{ex} \text{ or } \nodeSet^{qu},
\end{split}
\end{equation}

%\shixun{shall we add subscript of 'x' to the right side of equation 7?}==>Done
%in $\labelSet=\{\labelEmbedding_i\}_{i=1}^{m}$
\noindent where $W\in \mathcal{R}^{3d\times d}$ is a learnable weight matrix and $b$ is the learnable bias.
A common practice is to initialize each label node with a random vector~\citep{prodigy}. However, without contextual information, the resulting embeddings remain difficult to differentiate, even with prolonged training. Instead, we initialize the corresponding label node $\labelEmbedding_c$ by averaging the embeddings of examples belonging to the same class as below. This context-aware initialization potentially positions the label embedding nearer to its optimal representation, leading to more efficient training.
\begin{equation}
    \labelEmbedding_c \;=\; \frac{1}{|\mathcal{S}^{ex}_c |}\,\sum_{x \in \mathcal{S}^{ex}_c} \embedding^{ex}_x.
\end{equation}

\subsection{Discriminative Prompt Graph Learning}
\label{sec:learning}
The prompt graph is constructed by linking these initialized example nodes, query nodes, and label nodes as shown in Figure~\ref{fig:prompt_pipeline}. From the \emph{prompt graph} perspective, example node embeddings \(\embedding^{ex}\) provides contextual information for label node embedding \(\labelEmbedding\). Finally the example set $\mathcal{S}^{ex}$ and query set $\mathcal{S}^{qu}$ for label $c$ can be refined by an attention-based \(\GNNref\)~\citep{GAT}, with edge labels modulating the attention weights. 
%revise: what we refined is not set is each embedding

The task for training is to predict the label of each query node. To this end, classification logits $\logits_{x,c}$ for input $x$ and label $c$ are computed as the cosine similarity between each query embedding $\embedding^{qu}_{x}$ and label embedding $\labelEmbedding_{c}$: 
\begin{equation}
\logits_{x,c}
= \mathrm{cos\_sim}(\embedding^{qu}_x,\;\labelEmbedding_{c_x}),  \;c\in\class.
\end{equation}

To encourage each \(\labelEmbedding\) to be closer to embeddings of its own class $c$ and more distant from those of other classes, the model minimizes the Cross-Entropy (CE) loss $\lossCE$:
\begin{equation}
\lossCE=-\sum_{x}\log\frac{\exp\! \bigl(\logits_{x,c}\bigr)}{
  \displaystyle\sum_{c_i\in\class}\exp\!\bigl(\logits_{x,c_i}\bigr)}.
\end{equation}

% why we need auxiliary loss? CE usually be enough in commen sinario, but in this setting the labels come from agragation, and GNN have the inheritive intension to capture the shared charactors so the embedding generate through 2 GNN models can become very similar to each other. 

While the generated labels align the pre-train and downstream dataset~\citep{prodigy}, those labels lack inherent features, and the tendency of GNNs to capture shared structures causes them to become similar~\citep{bo2021beyond}, making the CE objective alone insufficient to enforce clear inter‑class margins.
% How to achieve this? we dot product shows the 
To prevent label embedding collapse, we introduce an orthogonal loss $\lossOT$ alongside CE. Orthogonal loss $\lossOT$ explicitly discourages alignment between label embeddings. We normalize each label embedding $\labelEmbedding_{c_i}$ to $\ell_{c_i}$, then compute its dot product with all others—large projections indicate near-parallel directions and potential class overlap. $\lossOT$ penalizes these projections, pushing label embeddings toward mutual orthogonality while preserving their individual norms, as shown in Figure~\ref{fig:technique3}.
\begin{equation}
\label{orthLoss}
\lossOT \;=\; \sum_{i \neq j} 
        \bigl( \ell_{c_i}^{\top}\ell_{c_j} \bigr)^{2}.
\end{equation}

Taken together, our training objective combines the cross-entropy loss $\lossCE$ with orthogonal loss $\lossOT$, weighted by a balancing coefficient $\lambda$. Enable the model to learn label representations that are both well aligned with their queries and mutually distinctive, resulting in more discriminative and generalisable classification. We retain the attribute prediction loss as an auxiliary objective. During \textsc{MaskNode} augmentation on each augmented graph $\widetilde{\graph^{con}_{x}}$, we mask a subset of node attributes $F_v$. An MLP takes the learned node embeddings $E_v$ to predict the masked attributes, and we add a mean squared error (MSE) reconstruction loss as an auxiliary augmentation term:
\begin{equation}
\label{eq:attr-loss}
\mathcal{L}_{\text{attr}}(\widetilde{\graph^{con}_{x}})=\frac{1}{\lvert \mathcal{V}_x \rvert}\sum_{v \in \mathcal{V}_x}
\mathrm{MSE}\!\left(F_v,\, \mathrm{MLP}(E_v)\right),
\end{equation}
\noindent where $\mathcal{V}^D$ denotes the index set of nodes whose attributes are masked by \textsc{MaskNode}.

The loss sums the squared similarities of every label pair, and minimising $\lossOT$ drives these similarities toward zero, forcing different label embeddings toward mutual orthogonality and thereby preserving clear inter-class margins that the cosine-based CE loss alone cannot guarantee.
\begin{equation}
    \loss = \lossCE + \lambda\lossOT + \mathcal{L}_{\text{attr}}(\widetilde{\graph^{con}_{x}}).
\end{equation}

% \end{document}

% \documentclass[main_anonymous.tex]{subfiles}   % 路径指向主文件
% \begin{document}

\section{Experiment}
\subsection{Experimental Setup}

\smallskip
\noindent \textbf{Evaluation Tasks}.
We conduct experiments on two types of tasks: \emph{node classification} and \emph{link prediction}. The node classification task is performed on homogeneous citation graphs, while the link prediction task is carried out on a heterogeneous knowledge graphs. To evaluate the cross-dataset inductive ability, we construct experiments with separate training and downstream datasets for both tasks. Unlike settings where training and evaluation are performed on different splits of the same graph, our training and downstream datasets are entirely distinct graphs, each with its own structure, label space, and feature distribution.
All training is conducted in a self-supervised few-shot setting. The training task is formulated as an $m$-way classification problem to identify the class of each input among $m$ candidate classes. During sample collection, we sample 3 examples and 4 queries for each class (ways) from the training dataset to construct the support and query sets, respectively. Training is then performed using these sets. The model is frozen after training, and no parameter updates are performed during downstream. In downstream evaluation, to simulate realistic deployment scenarios with limited labeled data, we provide $k$ labeled examples per class (way) all drawn from the training split of each downstream datasets. Unless otherwise specified, $k = 3$ in all experiments. %{\color{red} mention $m-$way.}
In node classification, we sample nodes into example and query sets to predict query labels in the downstream dataset. In the link prediction task, we sample triples consisting of two nodes and the edge between them. The goal is to predict the edge type. 

\smallskip
\noindent \textbf{Datasets}.
For the \emph{node classification} task, we pre-train the model on a sampled subset of MAG240M~\citep{mag} and evaluate cross-graph transfer on arXiv~\citep{arxiv}. 
For the \emph{link prediction} task, we pre-train on Wiki~\citep{FB15K-237NELL} and test on three knowledge graphs: 
ConceptNet~\citep{conceptnet}, FB15K-237~\citep{FB15K-237NELL} and NELL~\citep{FB15K-237NELL}. This design allows us to examine both homogeneous graph transfer and heterogeneous transfer under a unified evaluation framework. Dataset statistics are summarized in Table~\ref{tab:dataset_stats}. We construct a subset of MAG240M mainly for computational feasibility and to better support an inductive evaluation setting in a more accessible hardware environment. The subset is generated by a controlled neighborhood-expansion procedure: starting from sampled seed nodes, we iteratively collect their local neighbors while bounding the hop range, the number of neighbors per hop, and the total number of nodes in each sampled subgraph. This preserves representative local structure and semantic context while keeping preprocessing and training costs manageable.

%apply a BFS-based sub-sampling strategy that starts from the highest-degree node and expands its neighborhoods until reaching 20 million nodes and 35 million edges
%Maybe put this in appandix?

\begin{table}[!htbp]
\centering
\caption{Dataset statistics}
\begin{tabular}{l r r r}
\hline
\textbf{Dataset} & \textbf{\# Nodes} & \textbf{\# Edges} & \textbf{\# Classes} \\
\hline
MAG240M     & 122M & 1.3B & 153 \\
Wiki        & 4.8M & 5.9M & 639 \\
arXiv       & 169K & 1.2M & 40  \\
ConceptNet  & 791K & 2.5M & 14  \\
FB15K-237   & 15K  & 268K & 200 \\
NELL        & 69K  & 181K & 291 \\
\hline
\end{tabular}
\label{tab:dataset_stats}
\end{table}

\begin{table*}[!h]  % 跨两栏用 table*
  \centering
    \caption{Node Classification Accuracy (±std) on Arxiv. This table shows the accuracy and standard deviation in arXiv, in 5 settings of \{3,5,10,20,40\} ways (number of classes). All trained on the same 20M-node subgraph, and evaluated on Arxiv.}
  \small
  \begin{tabular}{cccccccc}
    \toprule
    \textbf{Way} & \textbf{\ourmethod} & \textbf{PG-NM} & \textbf{Contrastive} & \textbf{Graph Prompter}& \textbf{All In One} & \textbf{GPPT}& \textbf{GraphPrompt} \\
    \midrule
    \textbf{3}  & \textbf{66.09 {\fontsize{8}{9}\selectfont(±0.08)}} & 56.82 {\fontsize{8}{9}\selectfont(±0.08)} & 56.01 {\fontsize{8}{9}\selectfont(±0.09)} & 48.40 {\fontsize{8}{9}\selectfont(±0.17)}& 25.16 {\fontsize{8}{9}\selectfont(±0.27)} & 30.53 {\fontsize{8}{9}\selectfont(±0.28)}& 42.31 {\fontsize{8}{9}\selectfont(±0.24)} \\
    \textbf{5}  & \textbf{52.61 {\fontsize{8}{9}\selectfont(±0.06)}} & 42.50 {\fontsize{8}{9}\selectfont(±0.06)} & 41.18 {\fontsize{8}{9}\selectfont(±0.07)} & 32.61 {\fontsize{8}{9}\selectfont(±0.12)} & 27.06 {\fontsize{8}{9}\selectfont(±0.11)} & 22.62 {\fontsize{8}{9}\selectfont(±0.13)}& 19.93 {\fontsize{8}{9}\selectfont(±0.08)} \\
    \textbf{10}  & \textbf{36.94 {\fontsize{8}{9}\selectfont(±0.04)}} &28.62 {\fontsize{8}{9}\selectfont(±0.04)} & 27.35 {\fontsize{8}{9}\selectfont(±0.04)} & 19.17 {\fontsize{8}{9}\selectfont(±0.07)} & 9.75 {\fontsize{8}{9}\selectfont(±0.05)} & 11.28 {\fontsize{8}{9}\selectfont(±0.04)}& 10.92 {\fontsize{8}{9}\selectfont(±0.07)} \\
    \textbf{20}  & \textbf{24.03 {\fontsize{8}{9}\selectfont(±0.02)}} & 18.34 {\fontsize{8}{9}\selectfont(±0.02)} & 17.39 {\fontsize{8}{9}\selectfont(±0.02)} & 11.89 {\fontsize{8}{9}\selectfont(±0.04)} & 7.91 {\fontsize{8}{9}\selectfont(±0.10)} & 4.97 {\fontsize{8}{9}\selectfont(±0.10)}& 4.55 {\fontsize{8}{9}\selectfont(±0.03)} \\
    \textbf{40}  & \textbf{16.37 {\fontsize{8}{9}\selectfont(±0.01)}} & 10.63 {\fontsize{8}{9}\selectfont(±0.01)} & 10.79 {\fontsize{8}{9}\selectfont(±0.01)} & 5.97 {\fontsize{8}{9}\selectfont(±0.02)} & 2.22 {\fontsize{8}{9}\selectfont(±0.03)} & 1.57 {\fontsize{8}{9}\selectfont(±0.01)}& 3.17 {\fontsize{8}{9}\selectfont(±0.03)} \\
    \bottomrule
  \end{tabular}

  \label{table1}
\end{table*}

\begin{table*}[!htbp]  % 跨两栏用 table*
  \centering
  \small
   \caption{Link Prediction Accuracy (±std) on ConceptNet, FB15K-237, and NELL. The table reports test accuracy(\%) and standard deviation of \ourmethod{} and 2 baselines, all trained on Wiki and evaluated on these 3 graphs.} 
  \begin{tabular}{cccccc}
\toprule
\textbf{Dataset} & \textbf{Way} & \textbf{\ourmethod} & \textbf{PG-NM} & \textbf{Contrastive} & \textbf{Graph Prompter}\\
\midrule
\multirow{1}{*}{\textbf{ConceptNet}}
  & \textbf{4}  & 46.3{\fontsize{8}{9}\selectfont(±0.14)} & \textbf{46.94{\fontsize{8}{9}\selectfont(±0.61)}} & 44.01{\fontsize{8}{9}\selectfont(±0.61)} & 44.06{\fontsize{8}{9}\selectfont(±0.15)}\\
\addlinespace[2pt]
\cmidrule(lr){1-6}
\multirow{4}{*}{\textbf{FB15K-237}}
  & \textbf{5}   & \textbf{83.66{\fontsize{8}{9}\selectfont(±0.11)}} & 80.35{\fontsize{8}{9}\selectfont(±0.57)} & 81.35{\fontsize{8}{9}\selectfont(±0.58)} & 82.34{\fontsize{8}{9}\selectfont(±0.07)}\\
  & \textbf{10}  & \textbf{73.23{\fontsize{8}{9}\selectfont(±0.10)}} & 71.68{\fontsize{8}{9}\selectfont(±0.45)} & 70.88{\fontsize{8}{9}\selectfont(±0.48)} & 50.71{\fontsize{8}{9}\selectfont(±0.06)}\\
  & \textbf{20}  & \textbf{62.91{\fontsize{8}{9}\selectfont(±0.07)}} & 59.90{\fontsize{8}{9}\selectfont(±0.35)} & 59.80{\fontsize{8}{9}\selectfont(±0.35)} & 35.74{\fontsize{8}{9}\selectfont(±0.04)}\\
  & \textbf{40}  & 47.96{\fontsize{8}{9}\selectfont(±0.05)} & 46.82{\fontsize{8}{9}\selectfont(±0.21)} & \textbf{49.39{\fontsize{8}{9}\selectfont(±0.23)}} & 27.63{\fontsize{8}{9}\selectfont(±0.03)}\\
\addlinespace[2pt]
\cmidrule(lr){1-6}
\multirow{4}{*}{\textbf{NELL}}
  & \textbf{5}   & 84.32{\fontsize{8}{9}\selectfont(±0.12)} & 82.39{\fontsize{8}{9}\selectfont(±0.53)} & 83.38{\fontsize{8}{9}\selectfont(±0.50)} & \textbf{89.8{\fontsize{8}{9}\selectfont(±0.06)}}\\
  & \textbf{10}  & \textbf{76.53{\fontsize{8}{9}\selectfont(±0.10)}} & 75.14{\fontsize{8}{9}\selectfont(±0.43)} & 74.54{\fontsize{8}{9}\selectfont(±0.46)} & 72.00{\fontsize{8}{9}\selectfont(±0.06)}\\
  & \textbf{20}  & \textbf{66.28{\fontsize{8}{9}\selectfont(±0.07)}} & 65.68{\fontsize{8}{9}\selectfont(±0.34)} & 65.68{\fontsize{8}{9}\selectfont(±0.34)} & 56.84{\fontsize{8}{9}\selectfont(±0.04)}\\
  & \textbf{40}  & 54.20{\fontsize{8}{9}\selectfont(±0.05)} & 54.91{\fontsize{8}{9}\selectfont(±0.22)} & \textbf{56.70{\fontsize{8}{9}\selectfont(±0.23)}} & 45.53{\fontsize{8}{9}\selectfont(±0.04)}\\
\bottomrule
\end{tabular}
  \label{table2}
\end{table*}

\smallskip
\noindent \textbf{Baselines}.
Given the limited research on tuning-free prompts that operate on both homogeneous and heterogeneous graphs, we first compare our method \textbf{CTP} with the only existing method of this kind, PRODIGY~\citep{prodigy}. Specifically, we adopt \textbf{PG-NM}, the self-supervised variant, as a baseline. We also include its strongest baseline, a \textbf{Contrastive} approach which leverages node IDs as pseudo-labels and applies data augmentation to construct positive and negative pairs, with inference performed via average-based nearest neighbors.
In addition, we also include four widely used prompt-based baselines: \textbf{All in One}~\citep{allinone}, \textbf{GPPT}~\citep{gppt},  \textbf{GraphPrompt}~\citep{graphprompt} and \textbf{Graph Prompter}~\citep{lv2025graphprompter}. To simulate realistic self-supervised scenarios, all methods are evaluated without post-pretraining prompt tuning or supervised signals for downstream tasks. We include all baselines in the node classification experiments. For link prediction, All-in-One, GPPT, and GraphPrompt are not considered, as these methods are not designed for heterogeneous graph scenarios.

% {\color{red} you need to mention which baselines will used in which task and why. since in the link predtion, many baselines are not included. people will ask why.}

\smallskip
\noindent \textbf{Parameter Settings}. In MAG240M, ArXiv, and Wiki, we extracted 2-hop neighborhood to capture richer structural and semantic context, while for ConceptNet, NELL, and FB15K-237, 1-hop neighborhood is sufficient due to their denser connectivity patterns. Our model architecture was implemented with an input feature dimension of 768 and an embedding dimension of 256. The GNN backbone consisted of one graph convolutional layer, followed by one meta-layer for higher-level representation learning. For optimization, we employed optimizer with a learning rate of $1\times10^{-3}$, weight decay of $1\times10^{-3}$, and dropout rate of 0. Each model was trained with a batch size of 5 for a total of 12 epochs. Regarding few-shot task configuration, we used a 3-way, 3-shot, 4-query setup during training, allowing the model to learn transferable representations under a constrained supervision setting. In each batch, we sampled 10 centroids to represent the local neighborhood prototypes. Unless otherwise specified, this configuration remained consistent across all experiments to ensure comparability among methods. %All algorithms, including the proposed method and baselines, were implemented under the same environment and parameter initialization. 

\smallskip

\noindent \textbf{Environments}. All experiments were conducted with a single NVIDIA RTX A5000 GPU (24 GB).  The complete implementation details, along with parameter definitions and scripts for data preprocessing, are provided in our codebase for full reproducibility. Our implementation is available at \url{https://zenodo.org/records/17290794}.
% {\color{red} use a paragraph to describe parameter setting. provide a code link at the end of this paragraph in the main paper. if there is no enough space, mention briefly here and refer to appendix. also remember mentioning embedding dimension size. }

% \subsubsection*{\textbf{Environments.}}
% All experiments were using a 128 GB DDR4 RAM, and a single NVIDIA RTX A5000 GPU (24 GB VRAM). 

\subsection{Results}
% \noindent\textbf{Results on Node Classification Tasks}
% As shown in Table~\ref{table1}, \ourmethod consistently achieves the highest accuracy across all settings. Notably, it outperforms the strongest baseline PG-NM by margins of +9.34\% (3-way), +10.03\% (5-way), and +6.41\% (10-way), while also maintaining the lowest standard deviation. These results confirm the model’s strong inductive capability in homogeneous graphs, especially under few-shot conditions.
\smallskip
\noindent \textbf{Exp 1 -- Results on Node Classification Tasks. }
Table~\ref{table1} shows that \ourmethod{} achieves the best accuracy across all settings.In comparison with the second-best performing baseline, PG-NM, it improves performance by \textbf{+9.27\%}, \textbf{+10.11\%}, \textbf{+8.32\%}, \textbf{+5.69\%}, and \textbf{+5.74\%} on 3-, 5-, 10-, 20-, and 40-way tasks, respectively. In particular, \ourmethod{} reaches \textbf{16.37\%} in the 40-way case, compared with \textbf{10.63\%} for PG-NM. It also maintains consistently low standard deviations, indicating strong stability across different difficulty levels. Among the baselines, the Contrastive approach remains competitive with PG-NM and even slightly outperforms it in the 40-way setting. Moreover, we observe that these baselines exhibit a more pronounced performance drop as the number of classes increases, underscoring the advantage of our method in more challenging scenarios.

\begin{table}[!h]
      \centering
            \caption{Ablation study of the three optimization components, example and query collection (O1), context construction (O2), and the graph learning objective (O3), on Arxiv.}
      \small
      \begin{adjustbox}{max width=\linewidth,center}
      \begin{tabular}{c|ccccc}
        \toprule
        \textbf{Way} & \textbf{PG-NM} & \textbf{O1} & \textbf{O1+O2} & \textbf{O1+O3} & \textbf{O1+O2+O3} \\
        \midrule
        \textbf{3} & 56.82 {\fontsize{8}{9}\selectfont(±0.08)} & 61.84 {\fontsize{8}{9}\selectfont(±0.08)} & 61.47 {\fontsize{8}{9}\selectfont(±0.08)} & 60.20{\fontsize{8}{9}\selectfont(±0.08)} & \textbf{66.16 {\fontsize{8}{9}\selectfont(±0.08)}} \\
        \textbf{5} & 42.50 {\fontsize{8}{9}\selectfont(±0.06)} & 47.59 {\fontsize{8}{9}\selectfont(±0.06)} & 48.09 {\fontsize{8}{9}\selectfont(±0.06)} & 46.63{\fontsize{8}{9}\selectfont(±0.06)}  & \textbf{52.53 {\fontsize{8}{9}\selectfont(±0.06)}} \\
        \textbf{10} & 28.62 {\fontsize{8}{9}\selectfont(±0.04)} & 31.96 {\fontsize{8}{9}\selectfont(±0.04)} & 32.69 {\fontsize{8}{9}\selectfont(±0.04)} & 32.23{\fontsize{8}{9}\selectfont(±0.04)}  & \textbf{36.93 {\fontsize{8}{9}\selectfont(±0.04)}} \\
        \textbf{20} & 18.34 {\fontsize{8}{9}\selectfont(±0.02)} & 20.90 {\fontsize{8}{9}\selectfont(±0.02)} & 21.63 {\fontsize{8}{9}\selectfont(±0.02)} & 21.53{\fontsize{8}{9}\selectfont(±0.02)}  & \textbf{24.03 {\fontsize{8}{9}\selectfont(±0.02)}} \\
        \textbf{40} & 10.63 {\fontsize{8}{9}\selectfont(±0.01)} & 14.18 {\fontsize{8}{9}\selectfont(±0.01)} & 13.61 {\fontsize{8}{9}\selectfont(±0.01)} & 13.42{\fontsize{8}{9}\selectfont(±0.01)} & \textbf{16.37 {\fontsize{8}{9}\selectfont(±0.01)}} \\
        \bottomrule
\end{tabular}

      \end{adjustbox}

      \label{table3}
    \end{table}

\smallskip
\noindent \textbf{Exp 2 -- Results on Link Prediction Tasks}.
We further evaluate link prediction, training on Wiki and testing on ConceptNet, FB15K-237, and NELL. This setting involves heterogeneous graphs and cross-graph transfer, making it substantially more challenging. As shown in Table~\ref{table2}, 
% \ourmethod{} achieves the best performance on most settings, with gains of \textbf{+3.31\%}, \textbf{+1.55\%}, and \textbf{+3.01\%} over PG-NM in the 5-, 10-, and 20-way tasks respectively on FB15K-237. On NELL, \ourmethod{} also outperforms PG-NM in the 5-, 10-, and 20-way settings, with improvements of \textbf{+1.93\%}, \textbf{+1.39\%}, and \textbf{+0.60\%}, respectively. Although it does not achieve the best result in every case, \ourmethod{} consistently remains among the top-performing methods while keeping variance lower than most settings. Overall, these results demonstrate that \ourmethod{} maintains strong and stable performance on heterogeneous graphs, while retaining good transferability across different knowledge graph benchmarks.
shows that \ourmethod{} consistently delivers strong and stable performance across all datasets and settings, achieving clear improvements over baselines in most cases. On FB15K-237, \ourmethod{} outperforms the next best method by \textbf{+3.31\%}, \textbf{+1.55\%}, and \textbf{+3.01\%} in 5-, 10-, and 20-way settings, respectively, and maintains smaller performance drops as task difficulty increases. On NELL, \ourmethod{} achieves gains of \textbf{+1.39\%} and \textbf{+0.60\%} in 10- and 20-way, while remaining competitive in other settings. On ConceptNet (4-way), \ourmethod{} still improves over Contrastive and Graph Prompter by \textbf{+2.29\%} and \textbf{+2.24\%}, respectively. Overall, these results demonstrate the effectiveness and strong generalization ability of \ourmethod{}.

 % {\color{red} this paragraph below is not associated with any result. better incorportate it into other paragraphs. Currently, the experiment analysis description is too short. we can make it longer now since we have one more page to use.}

\smallskip
\noindent \textbf{Exp 3 -- Ablation Study. }
Table~\ref{table3} presents an evaluation of the three optimization components: example and query collection (Section~\ref{sec:collection}), context construction (Section~\ref{sec:context}), and the graph learning objective (Section~\ref{sec:learning}), denoted as O1, O2, and O3, respectively. When a specific optimization is omitted (e.g., O2 in the combination O1+O3), the corresponding functionality is replaced by the native implementation~\citep{prodigy}. Each optimization component provides notable performance improvements when applied individually or in partial combinations. However, their combined application often results in a much more substantial gain, indicating that their effects are synergistic, with the greatest benefits emerging from their joint use.

\begin{figure*}[!htb]
    \centering
    \begin{subfigure}{0.19\textwidth}
        \includegraphics[width=\linewidth]{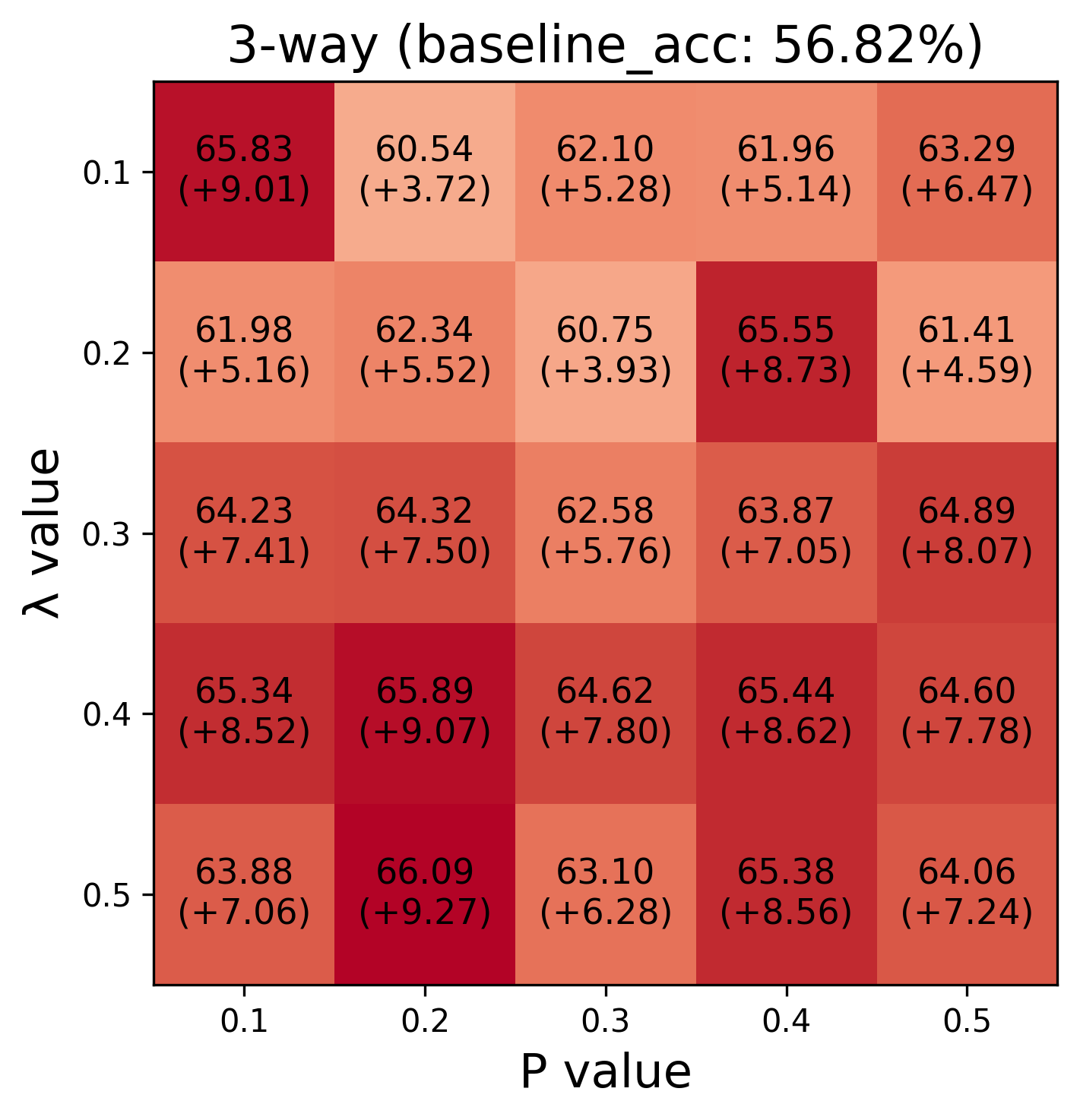}
        \caption{3-Ways}
    \end{subfigure}
    \begin{subfigure}{0.19\textwidth}
        \includegraphics[width=\linewidth]{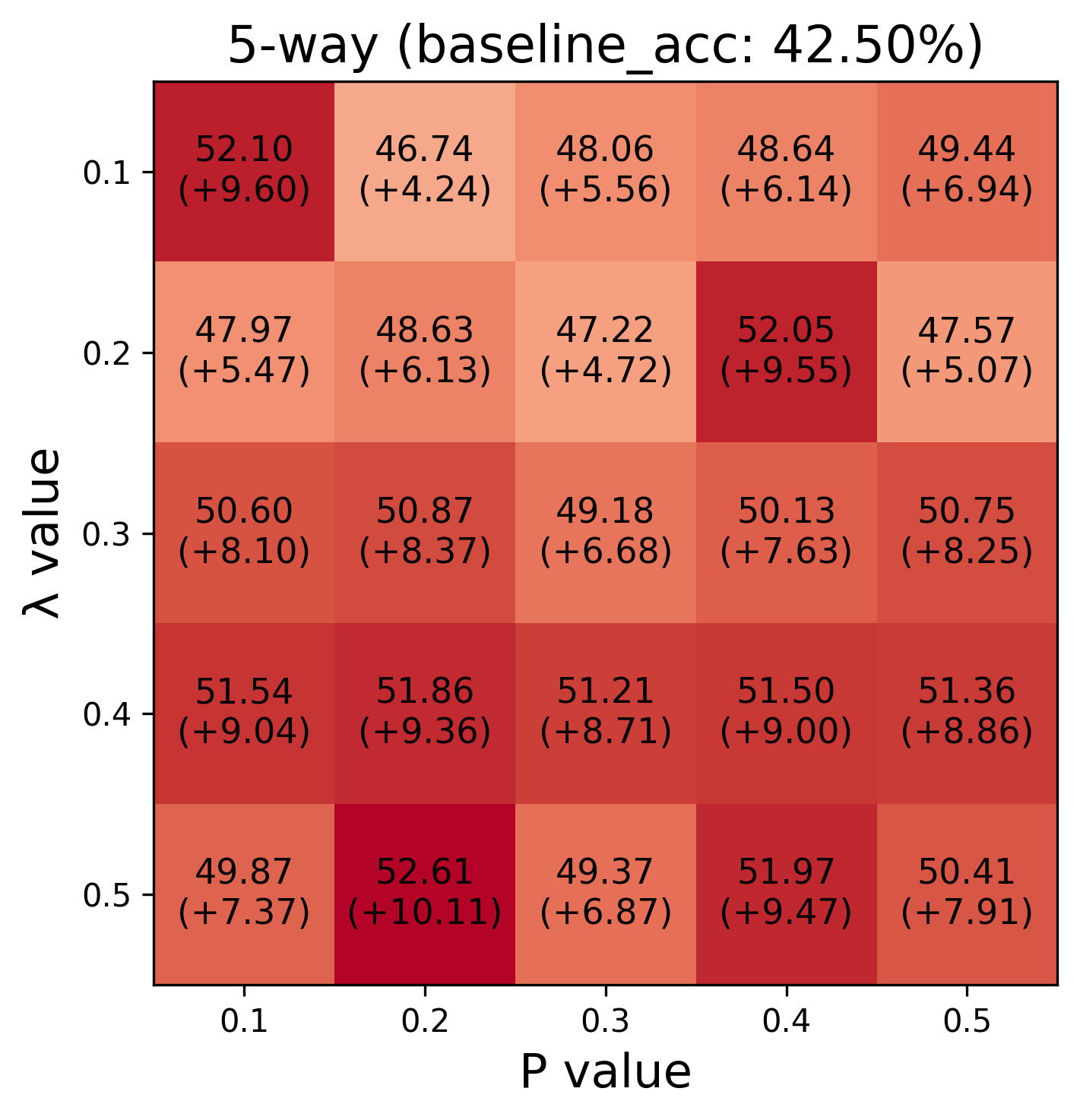}
        \caption{5-Ways}
    \end{subfigure}
    \begin{subfigure}{0.19\textwidth}
        \includegraphics[width=\linewidth]{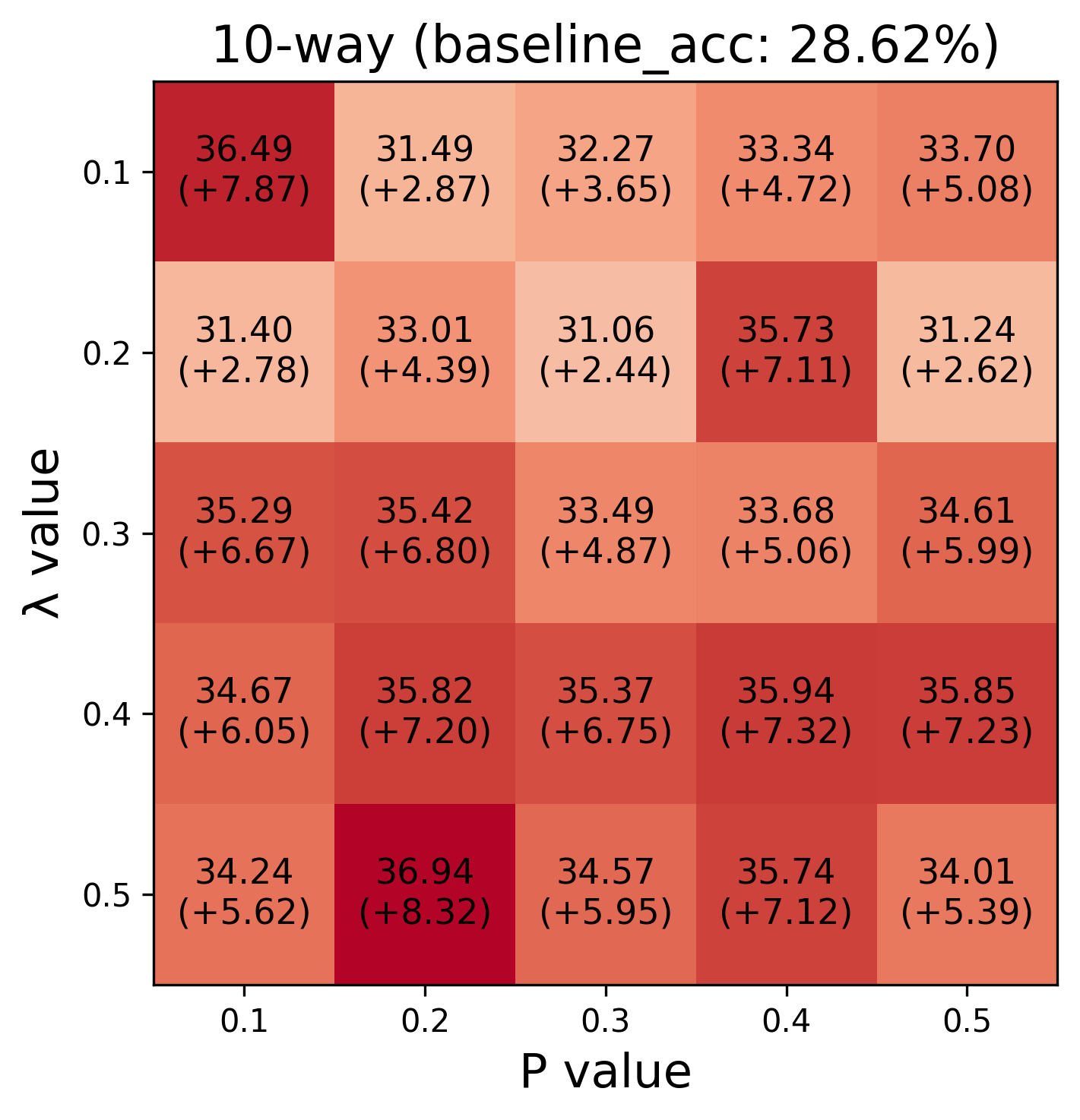}
        \caption{10-Ways}
    \end{subfigure}
    \begin{subfigure}{0.19\textwidth}
        \includegraphics[width=\linewidth]{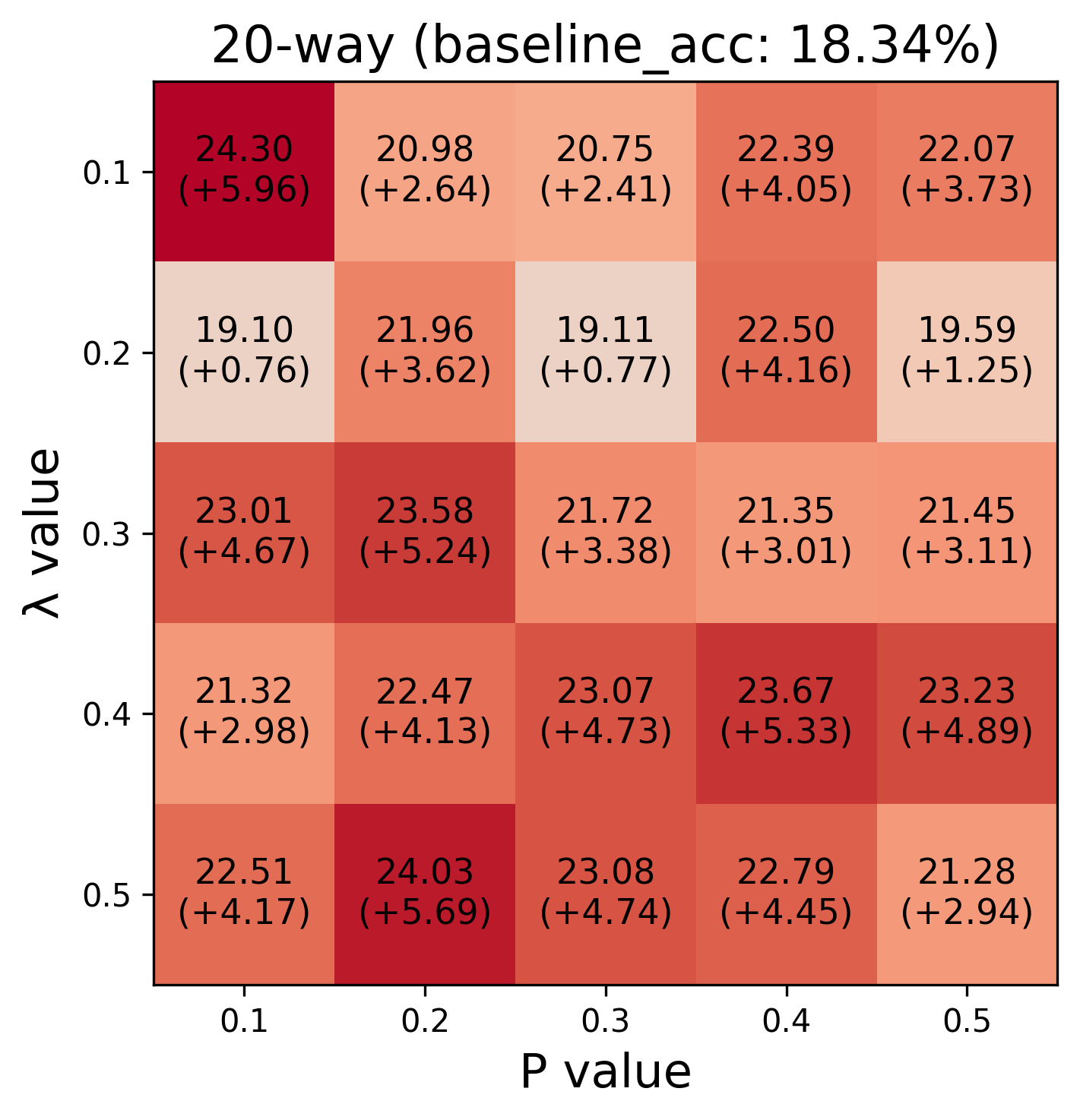}
        \caption{20-Ways}
    \end{subfigure}
    \begin{subfigure}{0.19\textwidth}
        \includegraphics[width=\linewidth]{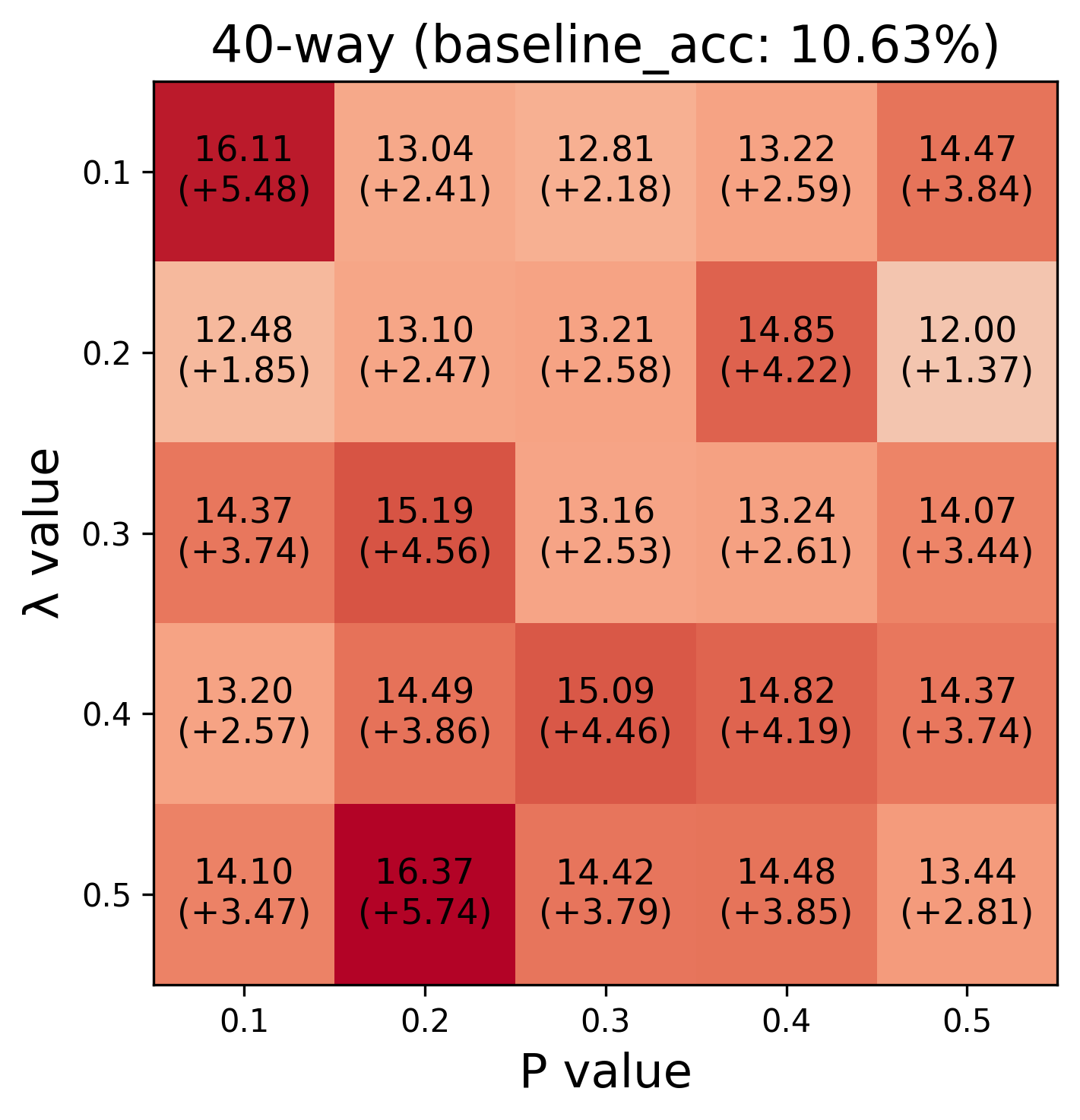}
        \caption{40-Ways}
    \end{subfigure}
    \caption{Parameter study: $\lambda$ and $p$. The heatmaps present the accuracy improvements of \ourmethod{} over PG-NM across different combinations of $\lambda$ (between $[0.1, 0.5]$) and $p$ (under 3-, 5-, 10-, 20-, and 40-way). In each cell, the top value reports the accuracy of \ourmethod{} for the corresponding parameter setting, while the value in parentheses shows the difference (i.e., \ourmethod{} accuracy minus PG-NM accuracy). Darker red shades indicate larger accuracy gains.}
    \label{fig:ways}
\end{figure*}
% \subsubsection{Exp4. Parameter Study}

\begin{figure*}[!htb]
    \centering
    \begin{subfigure}{0.195\textwidth}
        \includegraphics[width=\linewidth]{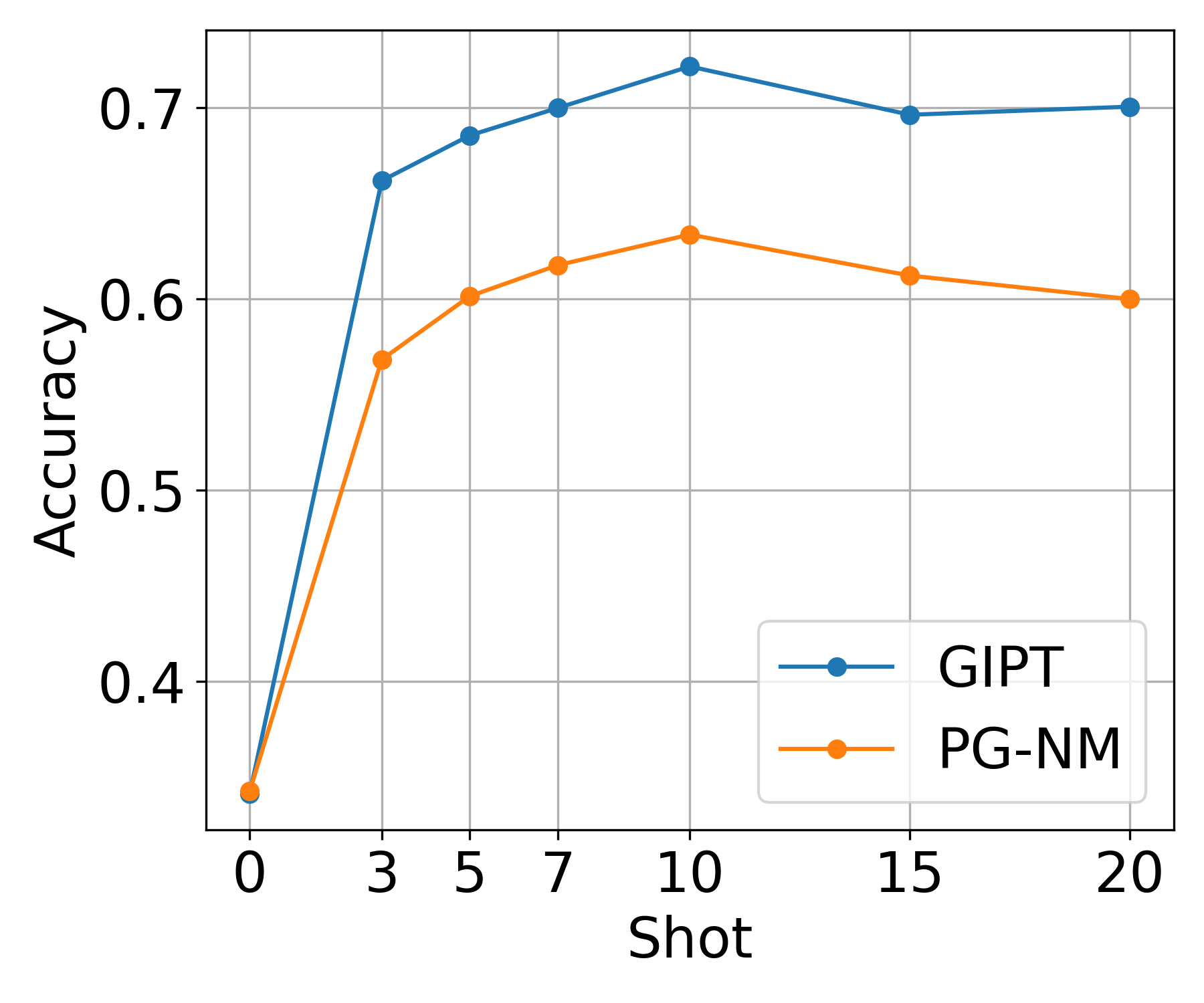}
        \caption{3-Ways}
    \end{subfigure}
    \begin{subfigure}{0.195\textwidth}
        \includegraphics[width=\linewidth]{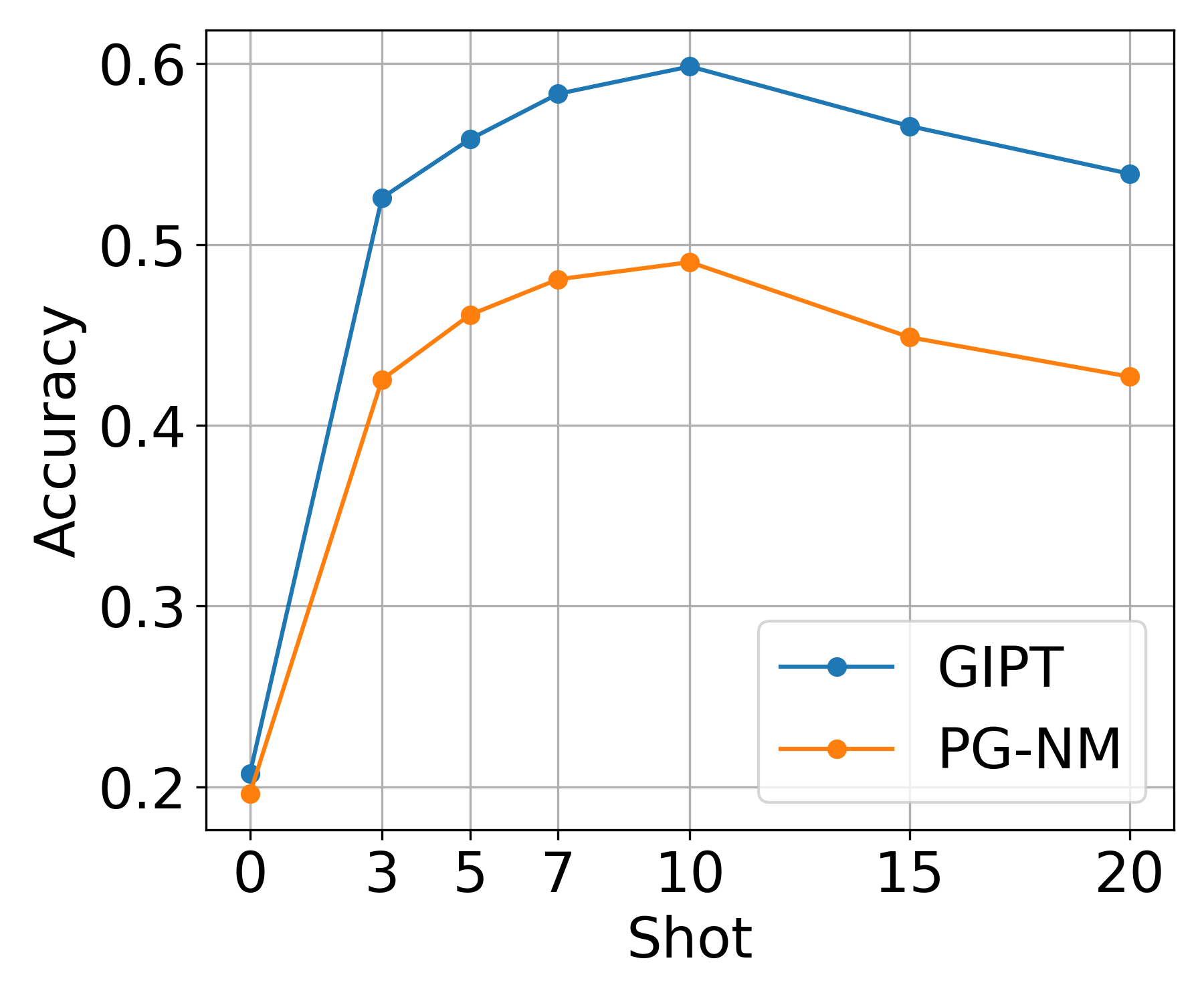}
        \caption{5-Ways}
    \end{subfigure}
    \begin{subfigure}{0.195\textwidth}
        \includegraphics[width=\linewidth]{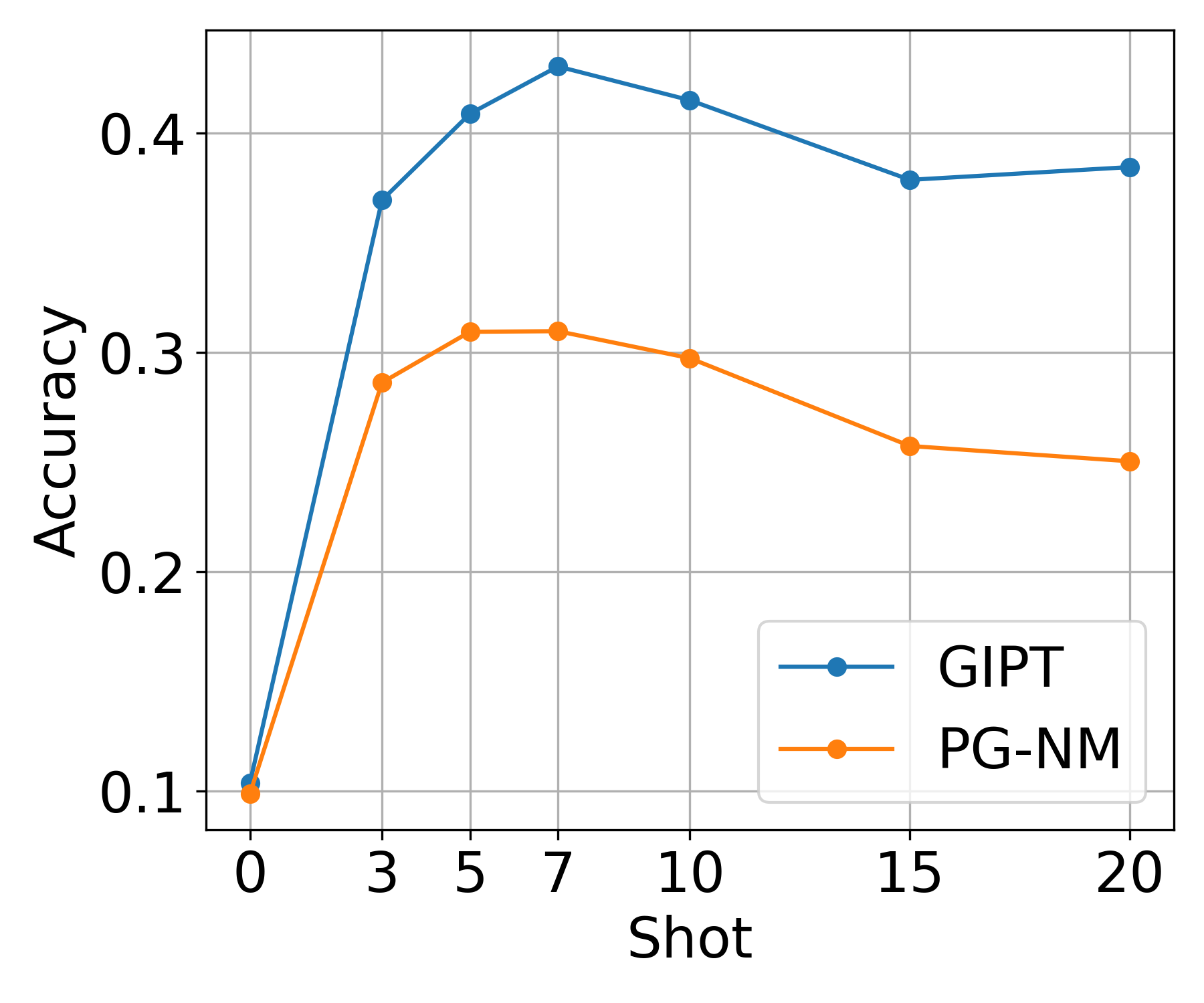}
        \caption{10-Ways}
    \end{subfigure}
    \begin{subfigure}{0.195\textwidth}
        \includegraphics[width=\linewidth]{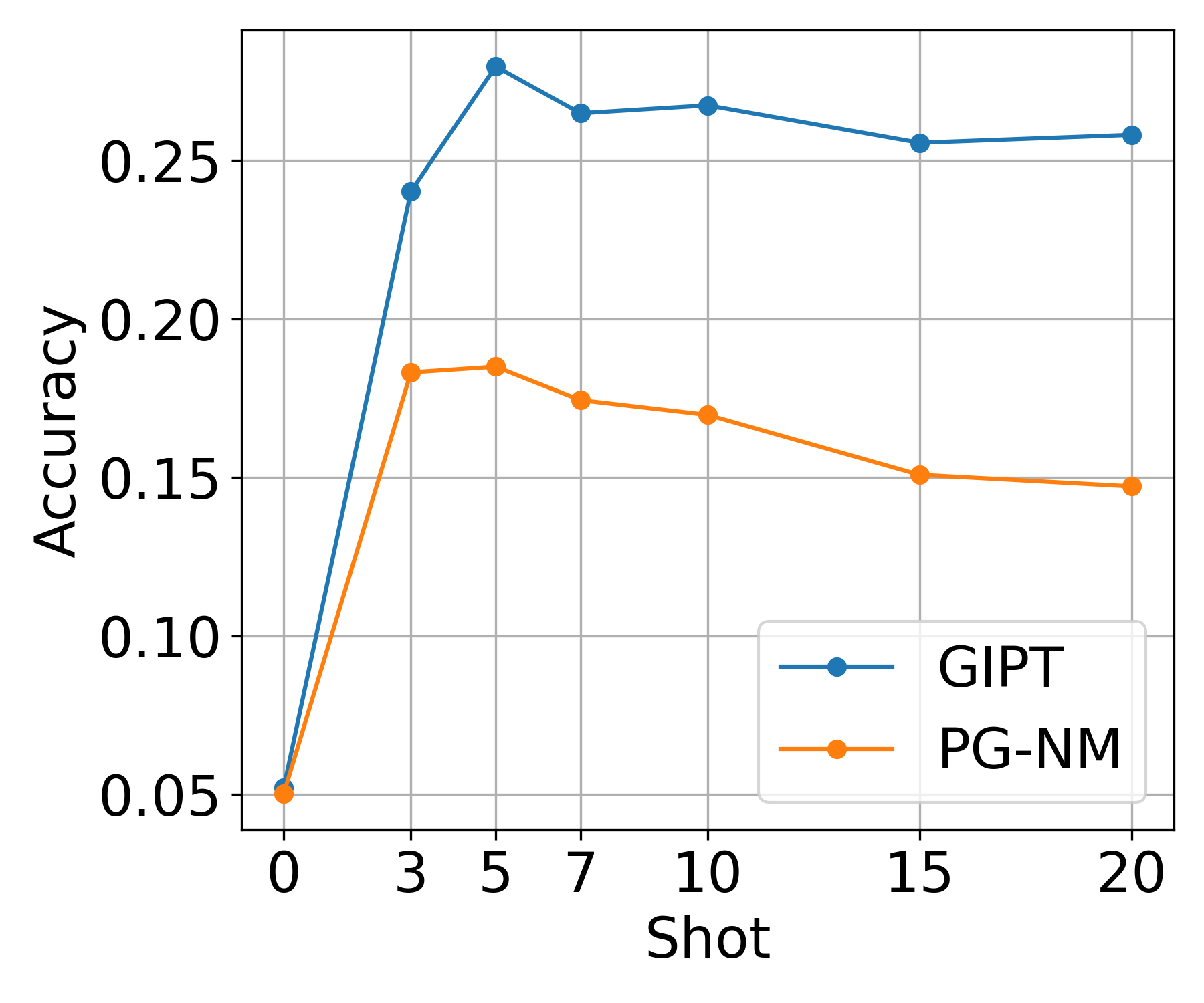}
        \caption{20-Ways}
    \end{subfigure}
    \begin{subfigure}{0.195\textwidth}
        \includegraphics[width=\linewidth]{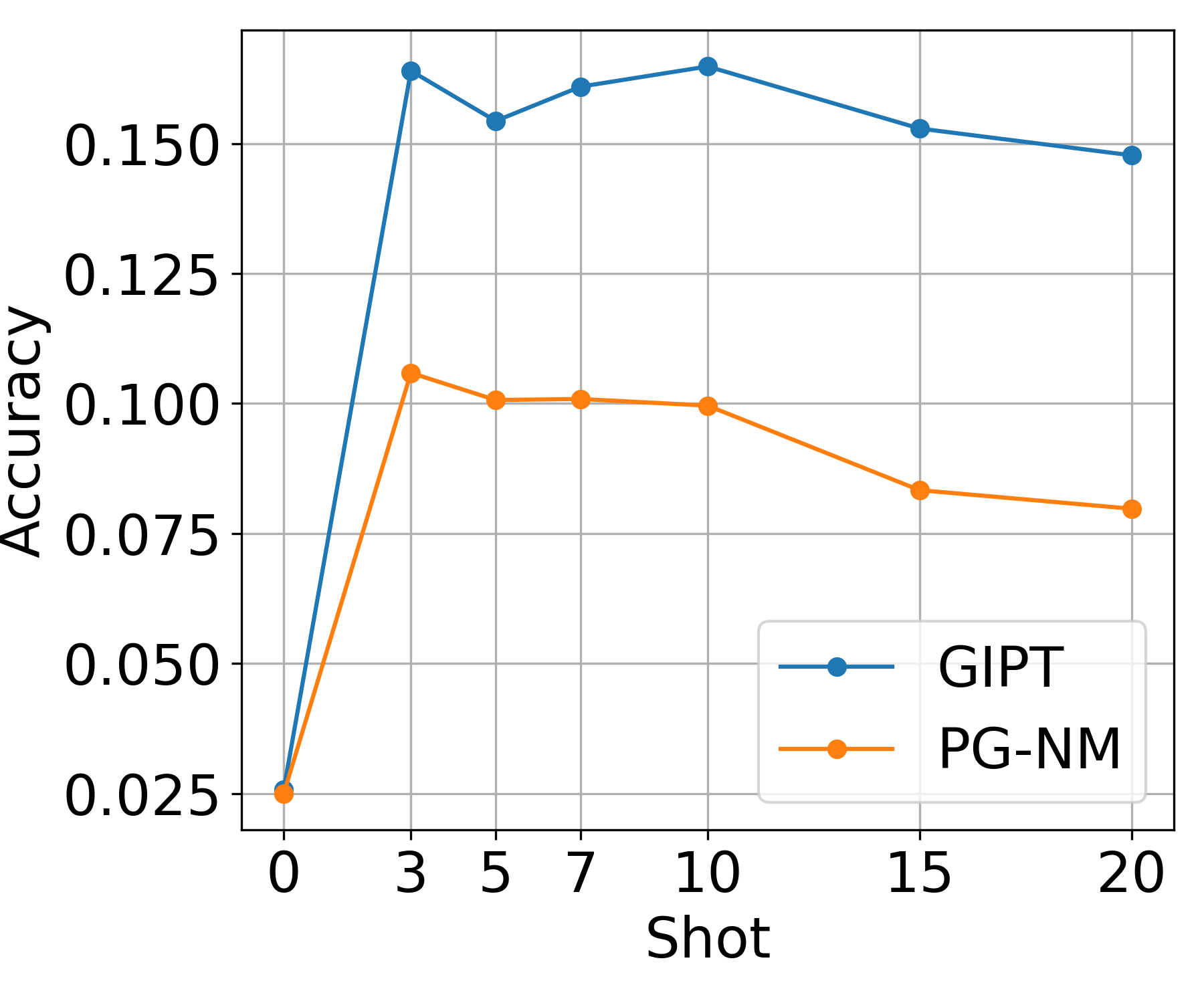}
        \caption{40-Ways}
    \end{subfigure}
    \caption{Parameter study: Number of Downstream Examples. Figures (a) - (e) show the results of varying the number of downstream examples $\{0, 3, 5, 7, 10, 15, 20\}$ across different way settings, with the blue line representing \ourmethod{} and the orange line representing PG-NM. As the number of ways increases and the task becomes more challenging, \ourmethod{} consistently achieves larger improvements over the baseline across all shot settings, demonstrating both superior and more stable performance. }
    \label{fig:shot_way_eval}
\end{figure*}
% As the number of ways increases and the task becomes more challenging, \ourmethod{} consistently achieves larger improvements over the baseline across all shot settings, demonstrating both superior and more stable performance.

\smallskip
\noindent \textbf{Exp 4 -- Parameter Study: $\lambda$ and $p$. }
During training, \ourmethod{} introduces two parameters: the orthogonal loss weight $\lambda$ and the protection ratio $p$ used in augmentation. We conduct experiments to examine their impact on performance, exploring values in the range $[0.1, 0.5]$ under all way settings using a node classification task, with training on MAG240M and evaluation downstream on Arxiv. Figure~\ref{fig:ways} illustrates the accuracy improvements of \ourmethod{} over PG-NM, where darker red shades indicate larger accuracy gains. The results show that \ourmethod{} outperforms PG-NM across all $\lambda$ and $p$ combinations and under every way setting, highlighting its robustness and stability against varying parameter choices.

\smallskip
\noindent \textbf{Exp 5 -- Parameter Study: Number of Examples. }
The parameter examined here is the number of examples provided in the downstream task. Varying the size of these example sets can influence performance. We evaluate across all way settings by providing the trained model with 0, 3, 5, 7, 10, 15, and 20 examples. As shown in Figure~\ref{fig:shot_way_eval}, \ourmethod{} consistently outperforms PG-NM across all configurations. The performance gap widens as the number of ways increases: while PG-NM degrades sharply, \ourmethod{} remains stable, confirming its robustness to task difficulty and limited supervision.

% \begin{figure*}[!t]
%     \centering
%     \includegraphics[width=1\linewidth]{heatmap_all_ways_delta.png}
%     \caption{\textbf{Parameter study of $\lambda$ and $p$:} This figure shows heat maps of $\lambda$ and $p$ combinations: the left for 3‑way and the right for 10‑way tasks. Each cell reports accuracy and improvement over baseline, with red for gains and blue for drops.}
%     \label{heatmap}
% \end{figure*}

\section{Conclusion and Future Work}
\noindent We present \ourmethod, one of the first tuning‑free GNN prompting frameworks. Pre‑trained once on a single source graph, \ourmethod{} can be directly applied to unseen graphs without parameter update, enabling seamless adaptation to both node‑ and edge‑level tasks. Extensive experiments show that \ourmethod{} achieves robust performance across diverse graph types and surpasses existing prompt‑based baselines in both accuracy and stability. Looking ahead, we aim to extend \ourmethod{} to graph-level and dynamic-graph tasks. These directions position \ourmethod{} as a general and adaptive framework for graph learning.

\section*{Acknowledgement}
\noindent This research did not receive any specific grant from funding agencies in the public, commercial, or not-for-profit sectors.

\clearpage %%Remove this from your manuscript

% % Figure
% \begin{figure}%[]
%   \centering
% %    \includegraphics{}
%     \caption{}\label{fig1}
% \end{figure}

% \begin{table}%[]
% \begin{tabular*}{\tblwidth}{@{}LL@{}}
% \toprule
% \midrule
%  & \\
%  & \\
%  & \\
%  & \\
% \bottomrule
% \end{tabular*}
% \end{table}

% Uncomment and use as the case may be
%\begin{theorem} 
%\end{theorem}

% Uncomment and use as the case may be
%\begin{lemma} 
%\end{lemma}


% \section{}\label{}


\bibliographystyle{apalike}

% % Loading bibliography database
\bibliography{cas-refs}
% \printbibliography

% \usepackage[backend=biber,style=authoryear]{biblatex}
% \addbibresource{cas-refs.bib}



\end{document}